\documentclass{article}

\PassOptionsToPackage{numbers, compress}{natbib}

\usepackage[preprint]{neurips_2022}

\usepackage[utf8]{inputenc} 
\usepackage[T1]{fontenc}    
\usepackage{hyperref}       
\usepackage{url}            
\usepackage{booktabs}       
\usepackage{amsfonts}       
\usepackage{nicefrac}       
\usepackage{microtype}      
\usepackage{xcolor}         

\usepackage{enumitem}
\usepackage{amssymb}
\usepackage{amsmath}
\usepackage{tikz}
\pgfdeclarelayer{nodelayer}
\pgfdeclarelayer{edgelayer}
\pgfsetlayers{edgelayer,nodelayer,main}
\usepackage{pgfplots}
\usepackage{caption}
\usepackage{subcaption}
\usepackage{soul}

\title{Optimal Decision Diagrams for Classification}

\author{%
  Alexandre M. Florio \\
CIRRELT \& SCALE-AI Chair in Data-Driven Supply Chains \\
Department of Mathematical and Industrial Engineering, Polytechnique Montr{\'e}al, Canada \\
  \texttt{aflorio@gmail.com} \\
  \And
  Pedro Martins \\
  Department of Computer Science,
  Pontifical Catholic University of Rio de Janeiro, Brazil \\
  \texttt{pmartins@inf.puc-rio.br} \\
  \And
  Maximilian Schiffer \\
  School of Management \& Munich Data Science Institute\\
  Technical University of Munich, Germany \\
  \texttt{schiffer@tum.de} \\
  \And
  Thiago Serra\\
  Freeman College of Management\\
  Bucknell University, USA\\
  \texttt{thiago.serra@bucknell.edu} \\
  \And
  Thibaut Vidal \\
 CIRRELT \& SCALE-AI Chair in Data-Driven Supply Chains \\
Department of Mathematical and Industrial Engineering, Polytechnique Montr{\'e}al, Canada \\
  Department of Computer Science, Pontifical Catholic University of Rio de Janeiro, Brazil \\
  \texttt{thibaut.vidal@polymtl.ca}
}

\newtheorem{theorem}{Theorem}

\newcommand{\cA}{\mathcal{A}}
\newcommand{\cG}{\mathcal{G}}
\newcommand{\cV}{\mathcal{V}}
\newcommand{\cVI}{\mathcal{V}^\textrm{\hspace*{0.025cm}I}} 
\newcommand{\cVC}{\mathcal{V}^\textrm{\hspace*{0.025cm}C}}
\newcommand{\cO}{\mathcal{O}}
\newcommand{\cC}{\mathcal{C}}

\begin{document}

\maketitle

\begin{abstract}
Decision diagrams for classification have some notable advantages over decision trees, as their internal connections can be determined at training time and their width is not bound to grow exponentially with their depth. Accordingly, decision diagrams are usually less prone to data fragmentation in internal nodes. However, the inherent complexity of training these classifiers acted as a long-standing barrier to their widespread adoption. In this context, we study the training of optimal decision diagrams (ODDs) from a mathematical programming perspective. We introduce a novel mixed-integer linear programming model for training and demonstrate its applicability for many datasets of practical importance. Further, we show how this model can be easily extended for fairness, parsimony, and stability notions. We present numerical analyses showing that our model allows training ODDs in short computational times, and that ODDs achieve better accuracy than optimal decision trees, while allowing for improved stability without significant accuracy losses.
\end{abstract}

\section{Introduction}\label{sec:introduction}
Decision diagrams, also known as decision graphs or decision streams, have a long history in logic synthesis and formal circuit verification \citep{Lee1959,Bryant1986,Bryant1992} as well as in optimization~\citep{Behle2007,Bergman2016,Lange2020} and artificial intelligence topics such as planning~\cite{Sanner2010,Castro2019}, knowledge compilation~\cite{Abio2012,Lai2013,Serra2020}, and constraint propagation~\cite{Andersen2007,Perez2015,Verhaeghe2018}. In machine learning, decision diagrams have recurrently emerged as a possible classification model \citep{Oliver1993,Oliveira1996,Mues2004,Shotton2013,Ignatov2018} or as a by-product of model compression algorithms applied on decision trees \citep{Breslow1997,Gossen2019,Choudhary2020}. A decision diagram for classification is represented as a rooted directed acyclic graph in which each internal node represents a splitting hyperplane, and each terminal node is uniquely associated to a class. The topology of the graph remains a free parameter of the model, such that decision diagram learning requires to \emph{jointly} determine the splitting hyperplanes and the node-connecting arcs.

Decision diagrams possess notable advantages over decision trees. Firstly, their width is not bound to grow exponentially with their depth, which allows training deep but narrow decision diagrams without quickly facing issues of data fragmentation  \citep{Shotton2013,Ignatov2018}. Moreover, additional degrees of freedom in their topology design permit to express a richer set of concepts and to achieve better model compression in memory-constrained computing environments \citep{Breslow1997,Kumar2017}. 

Despite these advantages, decision diagrams have been more rarely used than decision trees, as learning them remains inherently complex. A decision diagram topology cannot be easily optimized by construction or local optimization algorithms based on impurity measures. However, recent enhancements in global optimization techniques for decision tree training motivate us to reevaluate this issue. Indeed, optimal decision tree training through mathematical programming is becoming practical due to the formidable progress of hardware and mixed-integer linear programming solvers, which collectively led to speed-ups as high as $10^{11}$ between 1991 and 2015---most of which due to algorithmic improvements rather than hardware \citep{Bixby2012}. In view of this, we reevaluate the problem of searching for optimal decision diagrams (ODDs) through modern combinatorial optimization lenses and propose new mathematical models and efficient solution techniques to learn optimal decision diagrams. Specifically, our contributions are threefold:
\begin{enumerate}[nosep,leftmargin=*]
\item We propose the first mixed-integer linear program (MILP) to train decision diagrams for classification. This model effectively represents the decision diagram topology and the flow of samples within it, employing a limited number of binary variables. In practice, it includes exponentially fewer binary variables than \cite{Bertsimas2017} when applied to decision tree topologies. Furthermore, we include additional symmetry-breaking constraints that speed up the solution process, and provide efficient heuristic search strategies to obtain good primal solutions quickly.
\item We conduct an extensive computational study to evaluate our approach's scalability and compare the resulting decision diagrams with classical decision trees. We observe that training optimal decision diagrams requires a computational effort comparable to optimal decision-tree training but leads to more parsimonious and accurate models. 
\item As our MILP semantic permits to express various additional constraints with minimal adaptation, we discuss possible extensions to capture additional fairness, parsimony and stability requirements. We show the efficacy of such extensions in our numerical experiments for the stability case.
\end{enumerate}

\section{Related work}\label{sec:litt-review}
\paragraph{Optimal training of decision trees.} 
Standard construction algorithms for decision-tree training based on local impurity measures are not guaranteed to find the most accurate tree of a given size. To circumvent this issue, several works have been focused on global optimization algorithms. Optimal decision-tree training is known to be NP-hard \citep{Hyafil1976}, but solution methods for this problem went a long way from early dynamic programming algorithms \citep{Meisel1973,Payne1977} to modern solution approaches. Nowadays, these algorithms can find optimal trees for datasets with thousands of samples and hundreds of features \citep{Bertsimas2017,Demirovic2020,Firat2020}.

\citet{Carrizosa2020} recently conducted a comprehensive survey of mathematical programming approaches for optimal decision-tree training. Among the works surveyed, the paper of \citet{Bertsimas2017} represents a turning point, as it proposed a compact MILP formulation that could be solved within reasonable time for a wide range of datasets from the UCI machine learning repository. Subsequent methodological improvements occurred through sophisticated model reformulations and decomposition methods \citep{Aghaei2020,Firat2020}, permitting to achieve better linear programming bounds and quickly prune sub-optimal regions of the search space. Among other benefits, MILP-based approaches can handle combinatorial splits on categorical features, and they can easily include notions of fairness \citep{Aghaei2019,Ye2020} and parsimony \citep{Blanquero2020} through additional global objectives and constraints. Research in this field strives towards more effective solution approaches, based on combinations of branch-and-bound with dynamic programming \citep{Aglin2020,Demirovic2020}, constraint programming \citep{Verhaeghe2020}, or exploiting Boolean satisfiability (SAT) problem solvers \citep{Narodytska2018}. Some of these solution approaches also open new perspectives for other related training tasks, for example, in model compression \citep{Vidal2020a}.

\paragraph{Decision diagrams.}
Decision diagrams for classification have regularly reappeared in the machine learning domain. Early studies on the topic \citep{Oliver1993,Kohavi1995,Oliveira1996,akers1978} followed a minimum description length perspective and led to a first generation of learning algorithms -- often transforming an initial decision tree into a decision diagram. \citet{Oliver1993} argued that decision diagrams are particularly suitable to express disjunctive concepts (e.g., exclusive OR) and proposed an iterative node-merging algorithm. \citet{Kohavi1994} and \citet{Kohavi1995} first designed a bottom-up learning approach and then opted to train an oblivious decision tree and post-process it into a decision diagram through iterative merging. \citet{Oliveira1996} exploited efficient algorithms known from logic synthesis \citep{Bryant1986,Bryant1992} to manipulate ordered decision diagrams. They design an iterative greedy approach that merges nodes with similar sub-graphs to achieve a greater reduction of message length. An extension of support vector machines towards multi-class settings through decision diagrams was also presented in \citep{Platt2000}. Decision diagrams have also been used for model compression \citep{Breslow1997,Gossen2019} and as a surrogate model for neural networks in \citet{Chorowski2011}. They were generalized into ensembles called decision jungles in \citet{Shotton2013}. \citet{Ignatov2018} considered training deep decision diagrams, called decision streams, and reported a good performance on a range of datasets for credit scoring, aircraft control, and image classification. Recently, \citet{cabodi2021} studied binary decision diagrams in the context of interpretable machine learning, while \citet{hu2022optimizing} focused on optimizing binary decision diagrams via MaxSAT.

Most likely, the biggest obstacle towards the effective use of decision diagrams remains the ability to learn them efficiently. Indeed, most aforementioned learning algorithms first generate a decision tree and then transform it into a decision diagram. Unfortunately, these methods also often fix the order of the predicates through the tree, as this choice is known to lead to a difficult combinatorial optimization problem \citep{Bollig1996}. Accordingly, design decisions related to the graph topology (permitting to learn complex concepts) are ineffectively learned through trial and error. Our approach closes this significant methodological gap and, for the first time, allows to derive efficient algorithms for training decision diagrams.

\section{Mathematical formulation}\label{sec:methodology}
In this section, we mathematically formulate the ODD training problem as a MILP. We assume that we have a training dataset  \smash{$\{(\mathbf{x}^i,c^i)\}_{i=1}^n$} in which each $\mathbf{x}^i \in \mathbb{R}^d$ corresponds to a sample characterized by a $d$-dimensional feature vector and a class $c^i \in \mathcal{C}$. Our formulation takes as input the diagram's depth~$D$, i.e., the diagram's number of decision layers, and the width $w_l$ of each layer $l \in \{0,\dots,D-1\}$. This input constitutes the \emph{skeleton} of the decision diagram, and it guarantees that any final topology found during training (i.e., number of activated nodes per layer and their mutual connections) is contained in the skeleton. W.l.o.g., we can assume that $w_{l+1} \leq 2 w_l$ for all $l$. The MILP optimizes the decision diagram's topology and the splitting hyperplane of each internal node. We allow connections between internal nodes of consecutive layers and direct connections via \emph{long arcs} to terminal nodes representing the classes, as this permits to progressively assign a final classification for specific samples without necessarily passing through the complete decision diagram. Accordingly, the training algorithm can progressively send samples to the final leaves to fully exploit the remaining layers for classifying the other samples.

We designed our approach for numerical data, such that processing other data types requires prior transformation, e.g., one-hot encoding for categorical data, which is a common practice in decision-tree-based models. For numerical stability, we assume that each feature has been normalized within~$[0,1]$. Our model handles multiclass classification with a dedicated leaf for each class. It also naturally finds optimal multivariate (i.e., diagonal) splits without extra computational burden. We further show how it can be restricted to produce only univariate splits if necessary.

\subsection{Canonical formulation}
\begin{figure}[t]
\scalebox{0.8}
{
\tikzstyle{MyNode}=[fill=white, draw=black, shape=circle, minimum width = 0.5cm, minimum height = 0.5cm]
\tikzstyle{Connect}=[fill=black, draw=white, shape=circle, scale=0.7, ultra thick]
\tikzstyle{Arrow}=[-stealth]
\tikzstyle{ThinArrow}=[-stealth,thin]
\tikzstyle{ThickArrow}=[-stealth,very thick]
\tikzstyle{block} = [draw, rectangle, fill=gray!20, text centered, minimum height=8mm, node distance=10em]
\begin{tikzpicture}
	\begin{pgfonlayer}{nodelayer}
	    \draw [dashed,gray] (-4.7, 2.25) -- (4.7, 2.25);
	    \draw [dashed,gray] (-4.7, 0) -- (4.7, 0);
	    \draw [dashed,gray] (-4.7, -2.25) -- (4.7, -2.25);
	    \draw [dashed,gray] (-4.7, -4.5) -- (4.7, -4.5);
	    \node [style=MyNode] (0) at (0, 2.25) {$0$};
	    \node [style=Connect] (0a) at (-0.75, 1.5) {};
		\node [style=Connect] (0b) at (0.75, 1.5) {};
		\node [style=MyNode] (2) at (2, 0) {};
		\node [style=MyNode] (3) at (-2, 0) {$u$};
		\node [style=Connect] (4) at (-2.75, -0.75) {};
		\node [style=Connect] (5) at (-1.25, -0.75) {};
		\node [style=Connect] (6) at (1.25, -0.75) {};
		\node [style=Connect] (7) at (2.75, -0.75) {};
		\node [style=MyNode] (9) at (-4, -2.25) {$v$};
		\node [style=MyNode] (8) at (0, -2.25) {};
		\node [style=MyNode] (10) at (4, -2.25) {};
		\node [style=Connect] (11) at (-4.75, -3) {};
		\node [style=Connect] (12) at (-3.25, -3) {};
		\node [style=Connect] (13) at (-0.75, -3) {};
		\node [style=Connect] (14) at (0.75, -3) {};
		\node [style=Connect] (15) at (3.25, -3) {};
		\node [style=Connect] (16) at (4.75, -3) {};
		\node [style=MyNode, very thick, draw=white, fill=blue] (17) at (-2, -4.5) {};
		\node [style=MyNode, very thick, draw=white,  fill=green] (18) at (2, -4.5) {};
		\node at (-2, -4.9) {Class 1};
		\node at (2, -4.9) {Class 2};
		\node at (-1.1, -0.285) {$w^+_{iu}$};
		\node at (-2.9, -0.285) {$w^-_{iu}$};
		\node at (-3.75, -1.25) {$z^-_{iuv}$};
		\node at (5.3, 2.25) {{\color{gray}$\cVI_{0}$}};
		\node at (5.3, 0) {{\color{gray}$\cVI_{1}$}};
		\node at (5.3, -2.25) {{\color{gray}$\cVI_{2}$}};
		\node at (5.3, -4.5) {{\color{gray}$\cVC$}};
		\node at (-8, 1.125) [block, text width = 5cm]{\small Thick edges represent a possible decision-graph topology (selected by the training algorithm)};
		\node at (-8, -1.125) [block, text width = 5cm]{\small Flow variables \smash{$w^-_{iu}$, $w^+_{iu}$ and $z^-_{iuv}$} indicate the trajectory of sample~$i$. The following conditions always~hold:
		$(w^-_{iu} = 1) \Rightarrow (\mathbf{a}^\mathsf{T}_u x_i < b_u)$
		$(w^+_{iu} = 1) \Rightarrow (\mathbf{a}^\mathsf{T}_u x_i \geq b_u)$};
		\node at (-8, -3.375) [block, text width = 5cm]{\small The blue path corresponds to the possible trajectory of a sample classified as Class 1};
	\end{pgfonlayer}
	\begin{pgfonlayer}{edgelayer}
    	\draw [ThickArrow] (0) to (0a);
    	\draw [style=Arrow, blue, very thick] (0) to (0b);
    	\draw [ThinArrow] (0a) to (2);
    	\draw [style=Arrow,very thick] (0a) to (3);
    	\draw [style=Arrow,very thick, blue] (0b) to (2);
    	\draw [ThinArrow] (0b) to (3);
    	\draw [style=Arrow, blue, very thick] (2) to (6);
    	\draw [ThickArrow] (2) to (7);
    	\draw [ThickArrow] (3) to (4);
		\draw [ThickArrow] (3) to (5);
		\draw [ThinArrow] (4) to (8);
		\draw [ThickArrow] (4) to (9);
		\draw [ThinArrow] (4) to (10);
		\draw [ThickArrow] (5) to (8);
		\draw [ThinArrow] (5) to (9);
		\draw [ThinArrow] (5) to (10);
		\draw [ThinArrow] (6) to (8);
		\draw [ThickArrow, blue] (6) to (9);
		\draw [ThinArrow] (6) to (10);
		\draw [ThinArrow] (7) to (8);
		\draw [ThinArrow] (7) to (9);
		\draw [ThickArrow] (7) to (10);
		\draw [ThickArrow, blue] (9) to (11);
		\draw [ThickArrow] (9) to (12);
		\draw [ThickArrow] (8) to (13);
		\draw [ThickArrow] (8) to (14);
		\draw [ThickArrow] (10) to (15);
		\draw [ThickArrow] (10) to (16);
		\draw [ThickArrow, blue] (11) to (17);
		\draw [ThinArrow] (12) to (17);
		\draw [ThickArrow] (13) to (17);
		\draw [ThinArrow] (14) to (17);
		\draw [ThickArrow] (15) to (17);
		\draw [ThinArrow] (16) to (17);
		\draw [ThinArrow] (11) to (18);
		\draw [ThickArrow] (12) to (18);
		\draw [ThinArrow] (13) to (18);
		\draw [ThickArrow] (14) to (18);
		\draw [ThinArrow] (15) to (18);
		\draw [ThickArrow] (16) to (18);
	\end{pgfonlayer}
\end{tikzpicture}
}
\caption{Example of a graph $\cG$ with three layers of internal nodes ($w_1=2$ and $w_2=3$) and two terminal nodes. The thick edges indicate a possible decision diagram. The black connectors permit to illustrate flow-conservation within the graph. For clarity, the long arcs between the black connectors of layers $\cVI_{0}$ and $\cVI_{1}$ and the terminal nodes of $\cVC$ are not displayed.}
\label{fig:illustration-flow-vars}
\end{figure}
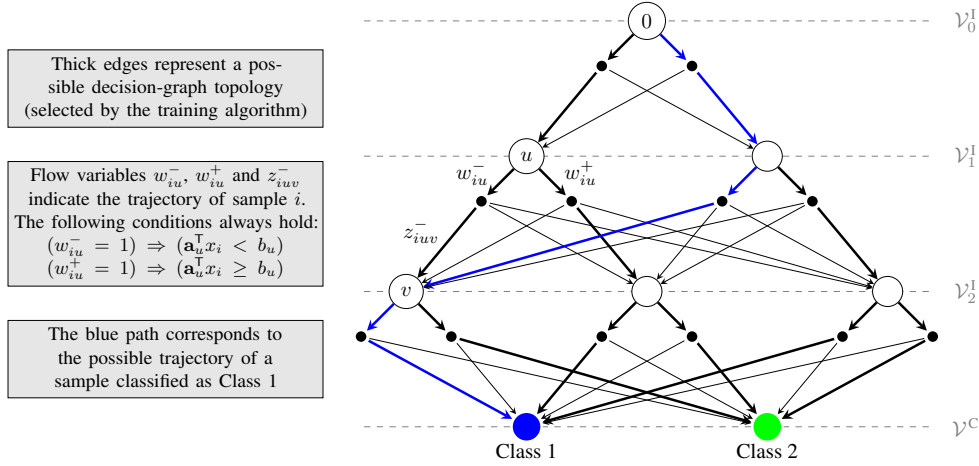

To represent the decision diagram, as illustrated on Figure~\ref{fig:illustration-flow-vars}, we define an acyclic graph $\cG = (\cV,\cA)$ with $\smash{\cV = \cVI \cup \cVC}$.
Each node $v \in \smash{\cVI}$ represents an internal node, and each node $v \in \smash{\cVC}$ represents a terminal node corresponding to a class $c_v$. 
We represent nodes by indices $\cV = \{0,\dots,|\cVI|+|\cVC|-1\}$; node $0 \in \cVI$ represents the root of the decision diagram and the remaining nodes are listed by increasing depth (from left to right on the figure). Let $\smash{\cVI_{l}}$ be the set of nodes at depth $l$, and let $\delta^-(v)$ and $\delta^+(v)$ be the predecessors and successors of each node $v \in \cV$. With these definitions, $\delta^-(0) = \varnothing$ and $\delta^+(0) = \smash{\cVI_{1} \cup \cVC}$. For $v \in \smash{\cVI_l}$ with $1 \leq l < D-1$, $\delta^-(v) = \smash{\cVI_{l-1}}$ and $\delta^+(v) = \smash{\cVI_{l+1} \cup \cVC}$. Finally, for $v \in \cVC$, $\smash{\delta^-(v) = \cVI}$ and $\delta^+(v) = \varnothing$. The decision diagram produced by our model will be a subgraph of $\cG$.

To formulate the training problem as a MILP, we start by defining the flow variables that represent the trajectory of the samples within the graph. We then connect these flows to the design variables defining the structure of the decision diagram and to those characterizing the splits.

\paragraph{Flow variables.} Each sample $i$ and internal node $u \in \cVI$ is associated to a pair of flow variables $w^-_{iu} \in [0,1]$ and $w^+_{iu} \in [0,1]$. A non-zero value in $w^-_{iu}$ (respectively $w^+_{iu}$) means that sample $i$ passes through node $u$ on the negative side of the separating hyperplane (on the positive side, respectively). Moreover, variables $z^-_{iuv} \in [0,1]$ (respectively $z^+_{iuv} \in [0,1]$) characterize the flow going from the negative and positive sides of $u$ to other nodes $v$. With these definitions, we can express flow conservation within the graph $\cG$ through the following conditions:
\begingroup
\allowdisplaybreaks
\begin{align}
&w^+_{iv} + w^-_{iv}  = 
\begin{cases}
1 & \text{if } v = 0 \\
\sum_{u \in \delta^-(v)} (z^+_{iuv} + z^-_{iuv}) & \text{otherwise} \\
\end{cases}
& v \in \cVI , i \in \{1, \ldots, n\} \label{start:formulation} \\
&w^-_{iu} = \sum_{v \in \delta^+(u)}  z^-_{iuv} &  u \in \cVI , i \in \{1, \ldots, n\} \label{myeq2} \\
&w^+_{iu} = \sum_{v \in \delta^+(u)}  z^+_{iuv} &  u \in \cVI, i \in \{1, \ldots, n\} \label{myeq3}
\end{align}
\endgroup

However, due to the interaction of the constraints coming from the hyperplanes (described later in this section), integrality of the flow variables is not guaranteed. To obtain integer sample flows, we add an additional binary variable $\lambda_{il} \in \{0,1\}$ for each sample $i \in \{1, \ldots, n\}$ and level $l \in \{0,\dots,D-1\}$, along with the following constraints:
\begin{align}
& \sum_{u \in \cVI_{l}} w^-_{iu}  \leq 1 - \lambda_{il} & l \in \{0,\dots,D-1\}, \  i \in \{1, \ldots, n\} \label{eq:my4} \\
& \sum_{u \in \cVI_{l}} w^+_{iu} \leq \lambda_{il} & l \in \{0,\dots,D-1\},  \  i \in \{1, \ldots, n\} \label{eq:my5}
\end{align}
With these constraints, sample $i$ can only go to the negative (respectively positive) side of any node $u$ of level $\smash{\cVI_{l}}$ if $\lambda_{il} = 0$ (respectively $\lambda_{il} = 1$). This allows us to use fewer binary variables compared to a direct definition of the $w^-_{iu}$ and $w^+_{iu}$ as binary (see Theorem~\ref{t2}), which is a desirable characteristic to allow for efficiently solving the MILP.

\paragraph{Decision diagram topology.}
We now connect the flow variables to the binary design variables that characterize the topology of the diagram. Here, we define one binary variable $d_u \in \{0,1\}$ for each $u \in \cV$ that takes value $1$ if this node is used in the classification (i.e., samples can pass through it). The terminal nodes and the root are always activated, so we impose $d_u = 1$ for $u \in 0 \cup \cVC$. For the negative and positive sides of each node $u \in \cVI$, we create binary design variables $y^-_{uv} \in \{0,1\}$ and $y^+_{uv} \in \{0,1\}$ taking value $1$ if and only if $u$ links towards $v$ on the negative and positive sides, respectively. The following constraints connect the design variables and the sample flows:
\begingroup
\allowdisplaybreaks
\begin{align}
&d_u = \sum_{v \in \delta^+(u)} y^+_{uv}      = \sum_{v \in \delta^+(u)} y^-_{uv} & u \in \cVI \label{myeq6} \\
& d_v \leq \sum_{u \in \delta^-(v)} (y^+_{uv} + y^-_{uv}) &  v \in \cVI - \{0\} \\
& y^+_{uv} + y^-_{uv} \leq d_{v} & u \in \cVI, v \in \delta^+(u) \\
&z^+_{iuv} \leq y^+_{uv}, & u \in \cVI, v \in \delta^+(u), i \in \{1, \ldots, n\} \label{myeq9}\\
&z^-_{iuv} \leq y^-_{uv} & u \in \cVI, v \in \delta^+(u), i \in \{1, \ldots, n\} \label{myeq10}
\end{align}
\endgroup
\paragraph{Symmetry-breaking constraints.}
Without any constraint for breaking symmetry, $2^{|\cVI|}$ equivalent topologies can be obtained by switching the positive- and negative-side arcs of each internal node and using opposite hyperplanes. Such symmetry has a dramatic negative impact on branch-and-bound-based MILP solution approaches. To circumvent this issue, we impose that arcs $(u,v)$ and $(u,w)$ such that $y^-_{uv} = 1$ and $y^+_{uw} = 1$ satisfy $v < w$ for each internal node $u \in \cVI$. This corresponds to the logical constraint $(y^-_{uv} = 1) \Rightarrow (y^+_{uw} = 0 \ \forall w \leq v)$, formulated as
\begin{align}
& y^-_{uv} + \sum_{w \in \delta^+(u), w \leq v} y^+_{uw} \leq 1 & u \in \cVI, v \in \delta^+(u)
\end{align}

To further reduce model symmetry and the number of equivalent topologies, we impose that the nodes along each layer must respect a weak in-degree ordering:
\begin{align}
& \sum_{w\in\delta^-(u)} (y^+_{wu}+y^-_{wu})\geq\sum_{w\in\delta^-(v)} (y^+_{wv}+y^-_{wv}) & l\in\{2,\dots,D-1\}, u,v\in\cVI_l, u<v
\end{align}

\paragraph{Linear separator variables and consistency with the sample flows.} We associate to each internal node $v \in \cVI$ a vector of variables $\mathbf{a}_v \in [-1,1]^d$ and a variable $b_v \in [-1,1]$ to characterize the splitting hyperplane. Samples $i \in \{1,\dots,n\}$ following the negative-side path should satisfy $\smash{\mathbf{a}^\mathsf{T}_v \mathbf{x}^i < b_v}$, whereas samples taking the positive-side path should satisfy $\smash{\mathbf{a}^\mathsf{T}_v \mathbf{x}^i \geq b_v}$. This is done by including \emph{indicator constraints} in our MILP that express the following implication logic:
\begin{align}
& (w^-_{iv} = 1) \Rightarrow (\mathbf{a}^\mathsf{T}_v \mathbf{x}^i+\varepsilon \leq b_v) &  i \in \{1,\dots,n\}, v \in \cVI \label{eq:logic1} \\
& (w^+_{iv} = 1) \Rightarrow (\mathbf{a}^\mathsf{T}_v \mathbf{x}^i  \geq b_v) &  i \in \{1,\dots,n\}, v \in \cVI \label{eq:logic2}
\end{align}

In Constraint~(\ref{eq:logic1}), $\varepsilon$ should be a small constant greater than the numerical precision of the solver (set to $\varepsilon = 10^{-4}$ in our experiments) that permits to express strict inequality. These logical constraints could be reformulated as linear constraints using a big-M transformation, leading to \smash{$\mathbf{a}^\mathsf{T}_v \mathbf{x}^i+\varepsilon \leq b_v + M(1-w^-_{iv})$} and \smash{$\mathbf{a}^\mathsf{T}_v \mathbf{x}^i \geq b_v - M(1-w^+_{iv})$}. However, as seen in \cite{Belotti2016}, modern MILP solvers generally benefit from directly specifying the implication logic and apply a big-M reformulation with tailored bound tightening.

Constraints~(\ref{eq:logic1}--\ref{eq:logic2}) represent general multivariate, i.e., oblique, splits. We can further restrict the model to produce univariate splits by adding for each feature $j \in \{1,\dots, d\}$ and internal node $v \in \cVI$ a binary variable $e_{vj} \in \{0,1\}$, along with constraints that impose the selection of a single feature:
\begin{align}
& \sum_{j = 1}^d e_{vj} = 1 & v \in \cVI \\
&  -e_{vj} \leq a_{jv} \leq e_{vj} & j \in \{1,\dots,d\}, v \in \cVI
\end{align}

\paragraph{Objective function.} In a similar fashion as in \cite{Bertsimas2017}, we optimize accuracy and an additional regularization term that favors simple decision diagrams with fewer internal nodes. The accuracy of the model can be computed by means of variables $w_{iv} \in [0,1]$ for each sample $i \in \{1,\dots,n\}$ and leaf $v \in \cVC$ expressing the amount of flow of $i$ reaching terminal node $v$ with class $c_v$. These variables must satisfy the following constraints:
\begin{align}
&w_{iv} = \sum_{u \in \delta^-(v)} (z^+_{iuv} + z^-_{iuv}) &  v \in \cVC,   i \in \{1, \ldots, n\}
\end{align}

With the help of these variables, we state our objective as
\begin{equation}
\min \frac{1}{n} \sum_{i = 1}^n \sum_{v \in \cVC} \phi_{iv} w_{iv} +  \frac{\alpha}{|\cVI|-1} \sum_{v \in \cVI - \{0\}} d_v,
\label{end:formulation}
\end{equation}
where $\phi_{iv}$ represents the mismatch penalty when assigning sample $i$ to terminal node $v$ (typically defined as $0$ if $c^i = c_v$ and $1$ otherwise), while $\alpha$ is a regularization parameter. The first term of the objective penalizes misclassified samples, whereas the second term penalizes complex models. With those conventions, the objective lies in $[0,1+\alpha]$ and an hypothetical value of $0$ would correspond to a model that achieves perfect classification with a single split at the root node.

\begin{theorem}
\emph{
Formulation (\ref{start:formulation}--\ref{end:formulation}) produces solutions in which all variables $\mathbf{w}$ and $\mathbf{z}$ take binary values, leading to a feasible and optimal decision diagram.
}
\label{t1}
\end{theorem}

\begin{theorem}
\emph{
Our formulation includes $\cO(n \log|\cV|)$ binary decision variables when applied to decision-tree skeletons.
}
\label{t2}
\end{theorem}

While Theorem~\ref{t1} establishes optimality for our MILP formulation, Theorem~\ref{t2} highlights its efficiency, illustrated in the context of decision tree skeletons where it improves upon the formulation of \cite{Bertsimas2017}, which requires $\cO(n |\cV|)$ binary decision variables. We give proofs to Theorems~\ref{t1} and~\ref{t2} in the supplemental material.

\subsection{Discussion}\label{sec:extensions}
Whereas classical training algorithms are often tailored to a specific setting and difficult to adapt to new requirements, our MILP approach for ODD training provides an extensible mathematical framework for expressing new requirements. We show how our MILP can be extended to fairness, parsimony, or stability measures. All extensions require minimal effort and only require extra linear constraints over the existing variables.

\paragraph{Fairness.} The confusion matrix can be directly calculated from the binary variables $w_{iv}$ of our model, which take value one if and only if sample $i$ reaches terminal node $v$ with class $c_v$. Hence, we can introduce classical fairness metrics into our MILP, either as an additional term in the objective or as a constraint, without significantly changing its complexity.
To illustrate this, consider a binary classification dataset in which outcome $1$ is the most desirable. For any subgroup~$g \subset \{1,\dots,n\}$, the number of samples classified as $1$ by the ODD can be calculated as \smash{$Z_g^\textsc{p} = \sum_{i \in g} w_{iv_\textsc{p}}$}, where $v_\textsc{p}$ is the terminal node associated to the positive class. This permits to express demographic parity \citep[see, e.g.,][]{Mehrabi2019} with the following linear constraints:
\begin{equation}
\sum_{i \in g_1} w_{iv_\textsc{p}} \geq \xi \sum_{i \in g_2} w_{iv_\textsc{p}},
\end{equation}
where $g_1$ and $g_2$ are two (non-necessarily disjoint) subgroups and $\xi$ represents a minimal discrepancy ratio between two subgroups. Compliance with the classic \emph{four-fifths} rule \citep{Biddle2006,Zafar2017} is achieved with~$\xi = 0.8$. In a similar fashion, the number of false-positives and false-negatives for any subgroup $g$ can be calculated as
\smash{$Z_g^\textsc{fp} = \sum_{i \in g, c^i \neq \textsc{p}} w_{iv_\textsc{p}}$} and
\smash{$Z_g^\textsc{fn} = \sum_{i \in g, c^i = \textsc{p}} (1-w_{iv_\textsc{p}})$}, permitting in turn to express most other classical fairness notions. e.g., equal opportunity or predictive equality \citep{Mehrabi2019}.

\paragraph{Parsimony and stability.}
Our MILP's flexibility for ODD training is not limited to fairness. It is possible to bound the number of activated nodes to a predefined limit $D$ to ensure parsimony by imposing
$
\sum_{i \in \cVI} d_i \leq D
$.
Finally, it is possible to impose that a minimum number $S$ of samples passes through each activated internal node to enhance stability, through the following conditions:
\begin{align}\label{eq:stability}
& \sum_{i=1}^n \sum_{u \in \delta^-(v)} (z^+_{iuv} + z^-_{iuv}) \geq S d_v & v \in \cVI.
\end{align}

\section{Training strategy}\label{sec:search-strategies}
We introduce a two-step search strategy, which permits to train an initial decision diagram quickly, and then refine it to eventually reach an optimal topology. First, we use an efficient multi-start construction and improvement heuristic to derive an initial topology. Then, we solve the complete MILP using an off-the-shelf branch-and-cut implementation (Gurobi in our case), where the value of the solution obtained in the first step is used as a cutoff value in the branch-and-bound search.

\paragraph{Step 1---Initial construction and improvement.}
We use a top-down construction approach that shares some common traits with CART \citep{Breiman1984a}. For each layer $l \in \{0,\dots,D-2\}$ and internal node $u \in \cVI_l$, we select the univariate split that maximizes the information gain. In contrast to CART, we determine the internal connections of the decision diagram at the training stage. Therefore, additional decisions need to be taken regarding the destination of the sample flows emerging from each layer. To this end, we adopt a greedy merging policy: as long as the number of sample flows is greater than $w_{l+1}$, we merge the pair of flows that least decreases the information gain. In the last layer connecting to the terminal nodes $\cVC$, we finally directly connect each sample flow to the terminal node of its most represented class.

To obtain a better initial diagram, we repeat the construction process for 60 seconds and consider only a random subset of $60\%$ of the features during each split selection. Additionally, we apply a bottom-up pruning strategy that consists of iteratively eliminating any internal (i.e., splitting) node that is (i) connected only to terminal nodes, and (ii) such that the removal improves Objective~\eqref{end:formulation}.

\paragraph{Step 2---Solution of the MILP.} In this step, we apply Gurobi on the complete MILP described in Section~\ref{sec:methodology}. Gurobi is a state-of-the-art solver for MILPs that utilizes a branch-and-cut process to derive lower and upper bounds on the objective value and to prune regions of the search space that have no chance to contain an optimal solution. To further guide the branch-and-bound search towards promising regions of the feasible space, we set an objective cutoff value equal to the value of the solution found in Step 1. The completion of this process gives a global optimum of the training model for the considered objective and regularization parameter. We set a CPU time limit of $T_\textsc{max} = 600$ seconds for this phase. The process terminates with the best solution found so far if $T_\textsc{max}$ is attained.

\section{Experimental analyses}\label{sec:experiments}
We focus our experiments on the same selection of 54 datasets as in \citet{Bertsimas2017}. All these datasets are publicly available from the UCI machine learning repository \citep{Dua2019}. They reflect a wide range of classification applications and contain between $47$ to $6435$ data points with $2$ to $484$ features. The list of datasets is accessible along with our detailed experimental results in the supplemental material. We split each dataset into a training, validation, and testing subset of samples with respective proportions of 50\%, 25\%, and 25\%. We repeat all our experiments five times for each dataset, using a different seed and thus a different random separation of the samples for each run.

All our algorithms have been implemented in Python 3.8 and can be readily executed from a single script. We use Gurobi 9.1.0 (via gurobipy) for solving the mathematical models. Each validation and test experiment has been run on a single thread of an Intel Gold 6148 Skylake @2.40GHz CPU. Overall, the experiments of this paper took eight hours on 10 CPUs of the mentioned type for five seeds. All data, source code, and additional details on the computational results are provided in the supplemental material and will be provided as a public Git repository upon publication.

We divide our computational experiments into two stages. Firstly, we conduct a calibration experiment using only the training and validation sets, considering different decision diagram skeletons and different levels of the regularization parameter~$\alpha$. The goal of this experiment is twofold: it permits to find the best skeleton and $\alpha$ hyperparameter for each dataset, and allows us to evaluate our methodology's computational performance, scalability, and sensitivity for different parameters. Finally, in the second stage of experiments, we use the test set to compare our optimal decision diagram models against the optimal decision tree model, using the best-found $\alpha$ hyperparameter for each model and dataset.

\subsection{Hyperparameters calibration and computational performance}\label{sec:performance}
We study the computational tractability of our approach for $\alpha\in\{0.01, 0.1, 0.2, 0.5, 1\}$ and for four decision diagram skeletons: (1--2--4--8), (1--2--4--4--4), (1--2--3--3--3--3--3) and (1--2--2--2--2--2--2--2), hereby referred to as Skeletons I to IV, respectively. All these skeletons have $15$ internal nodes; therefore, the final trained decision diagrams will contain no more than $15$ active nodes. We further employ Skeleton (1--2--4--8) to find optimal decision trees (ODTs). To this end, we specialize our model by fixing the decision variables representing the internal topology to match a decision tree; hence, generating optimal ODT solutions similar to \citet{Bertsimas2017}.

\paragraph{Computational performance.}
We run our algorithm for each dataset, split type (univariate and multivariate), random seed, skeleton, and value of $\alpha$. First, we evaluate our ability to find either the global optimum or an improved solution to the training problem, compared to the solution found in Step 1. Table~\ref{tab:performance} shows, for each skeleton and $\alpha$ combination, the number of runs (out of $54$ datasets $\times$ $5$ seeds $=270$ runs) for which a global optimum or an improved solution was found. Our algorithm can find optimal topologies in 32\% of all runs (4356 out of 13500 runs). Model difficulty is inversely proportional to $\alpha$ values, as large values of the regularization parameter discourages complex topologies with many active nodes. The difficulty of the training problem is generally sensitive to the number of samples in the dataset but relatively insensitive to the number of features (see results in the supplemental material). Because of the large size of some datasets and their resulting training models, in many cases optimality is not achieved within the short computational budget of 600 seconds used in our experiments. Still, Step 2 of our methodology improves upon the initial heuristic solution in 58\% of all runs, which demonstrates the value of the mathematical programming-based training approach and indicates that it might be possible to solve more instances to optimality when allocating larger computational budgets to each instance.

\begin{table}[t]
  \caption{Performance of the MILP-based training method}
  \label{tab:performance}
  \centering
\begin{tabular}{lrrrrrrrrrrr}
\toprule
 & \multicolumn{5}{c}{Proven optimality} & \multicolumn{5}{c}{Improved solution} & Total \\
\cmidrule(l){2-6}\cmidrule(l){7-11}
Skeleton & $\alpha=0.01$ & 0.1 & 0.2 & 0.5 & 1.0 & 0.01 & 0.1 & 0.2 & 0.5 & 1.0 \\
\midrule
I & 152 & 155 & 162 & 179 & 225 & 35 & 91 & 134 & 202 & 202 & 1537 \\
II & 150 & 154 & 162 & 179 & 227 & 36 & 91 & 124 & 194 & 193 & 1510 \\
III & 148 & 152 & 161 & 177 & 223 & 38 & 89 & 129 & 200 & 205 & 1522 \\
IV & 149 & 156 & 163 & 178 & 226 & 62 & 124 & 150 & 226 & 221 & 1655 \\
I (tree) & 155 & 158 & 163 & 177 & 225 & 30 & 98 & 131 & 208 & 207 & 1552 \\
\midrule
Total & 754 & 775 & 811 & 890 & 1126 & 201 & 493 & 668 & 1030 & 1028 & 7776 \\
\bottomrule
\end{tabular}
\end{table}

\subsection{Performance analyses and model comparison}
\paragraph{Accuracy of ODDs.}
We now compare the performance and structure of ODDs with those of ODTs. We use the best skeleton for each ODD, as obtained during the hyperparameter calibration phase of Section~5.1. We apply the same $\alpha$ hyperparameter calibration process for ODTs. To compare both methods based on near-optimal topologies, we focus on a subset of 18 datasets for which optimal ODDs and ODTs can be consistently obtained, and for which the final topologies are non-trivial (i.e., use at least two internal nodes). We refer the interested reader to the supplemental material for a more detailed discussion on the selection of those datasets. Moreover, we note that we do not extend this analysis to other heuristic tree models such as CART, since the superior performance of ODTs over CART has already been discussed for the same dataset collection in \citet{Bertsimas2017}.

Overall, ODDs and ODTs with univariate splits exhibit comparable accuracy as can be seen in the detailed numerical results in the supplemental material. With multivariate splits, however, ODDs show a higher accuracy compared to their ODT counterparts. Figure~\ref{fig:boxPerformance} shows the distribution of the classification accuracy on the test data over the selected datasets for the respective best ODT and ODD models, highlighting that the respective ODDs show fewer low-accuracy outliers. Their accuracy distribution is generally more compact and exhibits higher first, second, and third quartiles. Figure~\ref{fig:performanceSort} complements this analysis by representing the accuracy of all $18 \times 5$ instances for ODD and ODT in the multivariate splits case, sorted in ascending order. The sorted accuracy of the ODDs superposes that of the best ODTs, which points towards a better classification performance.

\begin{figure}[htbp]
	\begin{center}
		\begin{minipage}[t]{0.4\textwidth}
		\strut\vspace*{-\baselineskip}\newline
			\centering
%
%
\begin{tikzpicture}

\begin{axis}[%
width=3.5cm,
height=2.83cm,
at={(0cm,0cm)},
scale only axis,
xlabel near ticks,
ylabel near ticks,
unbounded coords=jump,
xmin=0.5,
xmax=2.5,
xtick={1,2},
xticklabels={{ODD},{ODT}},
ymin=0.4,
ymax=1.05,
ylabel={Test data accuracy},
axis background/.style={fill=white}
]
\addplot [color=black, dashed, forget plot]
  table[row sep=crcr]{%
1	0.877551020408163\\
1	1\\
};
\addplot [color=black, dashed, forget plot]
  table[row sep=crcr]{%
2	0.872881355932203\\
2	0.997084548104956\\
};
\addplot [color=black, dashed, forget plot]
  table[row sep=crcr]{%
1	0.522388059701492\\
1	0.72463768115942\\
};
\addplot [color=black, dashed, forget plot]
  table[row sep=crcr]{%
2	0.434782608695652\\
2	0.693877551020408\\
};
\addplot [color=black, forget plot]
  table[row sep=crcr]{%
0.925	1\\
1.075	1\\
};
\addplot [color=black, forget plot]
  table[row sep=crcr]{%
1.925	0.997084548104956\\
2.075	0.997084548104956\\
};
\addplot [color=black, forget plot]
  table[row sep=crcr]{%
0.925	0.522388059701492\\
1.075	0.522388059701492\\
};
\addplot [color=black, forget plot]
  table[row sep=crcr]{%
1.925	0.434782608695652\\
2.075	0.434782608695652\\
};
\addplot [color=blue, forget plot]
  table[row sep=crcr]{%
0.85	0.72463768115942\\
0.85	0.877551020408163\\
1.15	0.877551020408163\\
1.15	0.72463768115942\\
0.85	0.72463768115942\\
};
\addplot [color=blue, forget plot]
  table[row sep=crcr]{%
1.85	0.693877551020408\\
1.85	0.872881355932203\\
2.15	0.872881355932203\\
2.15	0.693877551020408\\
1.85	0.693877551020408\\
};
\addplot [color=red, forget plot]
  table[row sep=crcr]{%
0.85	0.797580395794681\\
1.15	0.797580395794681\\
};
\addplot [color=red, forget plot]
  table[row sep=crcr]{%
1.85	0.795918367346938\\
2.15	0.795918367346938\\
};
\addplot [color=black, only marks, mark=+, mark options={solid, draw=red}, forget plot]
  table[row sep=crcr]{%
nan	nan\\
};
\addplot [color=black, only marks, mark=+, mark options={solid, draw=red}, forget plot]
  table[row sep=crcr]{%
nan	nan\\
};
\end{axis}
\end{tikzpicture}%
			\vspace{0.45cm}
            \caption{Test data accuracy}\label{fig:boxPerformance}
		\end{minipage}%
		\hfill
		\begin{minipage}[t]{0.58\textwidth}
		\strut\vspace*{-\baselineskip}\newline
			\centering
%
%
\definecolor{mycolor1}{rgb}{0.00000,0.44700,0.74100}%
\definecolor{mycolor2}{rgb}{0.85000,0.32500,0.09800}%
\begin{tikzpicture}

\begin{axis}[%
width=6.5cm,
height=2.83cm,
at={(0cm,0cm)},
scale only axis,
xlabel near ticks,
ylabel near ticks,
xmin=0,
xmax=90,
xlabel={Dataset $\times$ run},
ymin=0.4,
ymax=1.05,
every axis plot/.append style={thick},
ylabel={Test data accuracy},
axis background/.style={fill=white},
legend style={at={(0.95,0.38)},legend cell align=left, align=left, draw=white!15!black}
]
\addplot[const plot, color=mycolor1] table[row sep=crcr] {%
1	0.434782608695652\\
2	0.5\\
3	0.5\\
4	0.5\\
5	0.513333333333333\\
6	0.543478260869565\\
7	0.567164179104477\\
8	0.597014925373134\\
9	0.611940298507462\\
10	0.623188405797101\\
11	0.623188405797101\\
12	0.625\\
13	0.625\\
14	0.625\\
15	0.626865671641791\\
16	0.634615384615384\\
17	0.64\\
18	0.652173913043478\\
19	0.652173913043478\\
20	0.653846153846153\\
21	0.653846153846153\\
22	0.692708333333333\\
23	0.693877551020408\\
24	0.701492537313432\\
25	0.701492537313432\\
26	0.710144927536231\\
27	0.711864406779661\\
28	0.71590909090909\\
29	0.720338983050847\\
30	0.723958333333333\\
31	0.72463768115942\\
32	0.731343283582089\\
33	0.731343283582089\\
34	0.739583333333333\\
35	0.746268656716417\\
36	0.75\\
37	0.759358288770053\\
38	0.760416666666666\\
39	0.761194029850746\\
40	0.764705882352941\\
41	0.764705882352941\\
42	0.770053475935828\\
43	0.78\\
44	0.791666666666666\\
45	0.795918367346938\\
46	0.795918367346938\\
47	0.796610169491525\\
48	0.803680981595092\\
49	0.806818181818181\\
50	0.807692307692307\\
51	0.8125\\
52	0.81283422459893\\
53	0.816326530612244\\
54	0.818181818181818\\
55	0.818181818181818\\
56	0.82\\
57	0.826923076923076\\
58	0.828220858895705\\
59	0.828220858895705\\
60	0.836734693877551\\
61	0.84090909090909\\
62	0.84090909090909\\
63	0.84090909090909\\
64	0.848484848484848\\
65	0.851851851851851\\
66	0.852272727272727\\
67	0.865030674846625\\
68	0.872881355932203\\
69	0.877300613496932\\
70	0.881355932203389\\
71	0.881481481481481\\
72	0.886363636363636\\
73	0.888888888888888\\
74	0.888888888888888\\
75	0.921052631578947\\
76	0.922600619195046\\
77	0.925696594427244\\
78	0.93188854489164\\
79	0.938080495356037\\
80	0.94074074074074\\
81	0.958333333333333\\
82	0.958333333333333\\
83	0.9625\\
84	0.9625\\
85	0.975\\
86	0.985422740524781\\
87	0.991253644314868\\
88	0.994169096209912\\
89	0.997084548104956\\
90	0.997084548104956\\
};
\addlegendentry{ODT}

\addplot[const plot, color=mycolor2] table[row sep=crcr] {%
1	0.522388059701492\\
2	0.543478260869565\\
3	0.565217391304347\\
4	0.58695652173913\\
5	0.594202898550724\\
6	0.623188405797101\\
7	0.625\\
8	0.625\\
9	0.626666666666666\\
10	0.633333333333333\\
11	0.641791044776119\\
12	0.646666666666666\\
13	0.652173913043478\\
14	0.652173913043478\\
15	0.653846153846153\\
16	0.656716417910447\\
17	0.656716417910447\\
18	0.681159420289855\\
19	0.695652173913043\\
20	0.708333333333333\\
21	0.711538461538461\\
22	0.716417910447761\\
23	0.72463768115942\\
24	0.729166666666666\\
25	0.746268656716417\\
26	0.746268656716417\\
27	0.746268656716417\\
28	0.75\\
29	0.75\\
30	0.75\\
31	0.755208333333333\\
32	0.755208333333333\\
33	0.760416666666666\\
34	0.764705882352941\\
35	0.769230769230769\\
36	0.770833333333333\\
37	0.775401069518716\\
38	0.776119402985074\\
39	0.776119402985074\\
40	0.779661016949152\\
41	0.791443850267379\\
42	0.793333333333333\\
43	0.793333333333333\\
44	0.795918367346938\\
45	0.795918367346938\\
46	0.799242424242424\\
47	0.805084745762711\\
48	0.8125\\
49	0.81283422459893\\
50	0.818181818181818\\
51	0.818181818181818\\
52	0.821969696969697\\
53	0.825757575757575\\
54	0.834355828220859\\
55	0.834355828220859\\
56	0.834355828220859\\
57	0.836734693877551\\
58	0.836734693877551\\
59	0.838983050847457\\
60	0.84090909090909\\
61	0.84090909090909\\
62	0.848484848484848\\
63	0.848484848484848\\
64	0.852760736196319\\
65	0.858895705521472\\
66	0.872881355932203\\
67	0.875\\
68	0.877551020408163\\
69	0.886363636363636\\
70	0.888888888888888\\
71	0.888888888888888\\
72	0.896296296296296\\
73	0.896296296296296\\
74	0.898305084745762\\
75	0.921052631578947\\
76	0.925696594427244\\
77	0.928792569659442\\
78	0.93188854489164\\
79	0.938080495356037\\
80	0.94074074074074\\
81	0.954166666666666\\
82	0.958333333333333\\
83	0.958333333333333\\
84	0.966666666666666\\
85	0.970833333333333\\
86	0.994169096209912\\
87	0.994169096209912\\
88	0.997084548104956\\
89	0.997084548104956\\
90	1\\
};
\addlegendentry{ODD}

\end{axis}
\end{tikzpicture}%
            \caption{Accuracy per dataset $\times$ run, ascending order}\label{fig:performanceSort}
		\end{minipage}
	\end{center}
\end{figure}

\begin{figure}[htbp]
    \centering
    \begin{subfigure}{0.4\textwidth}
    \centering
        \includegraphics[scale=0.135]{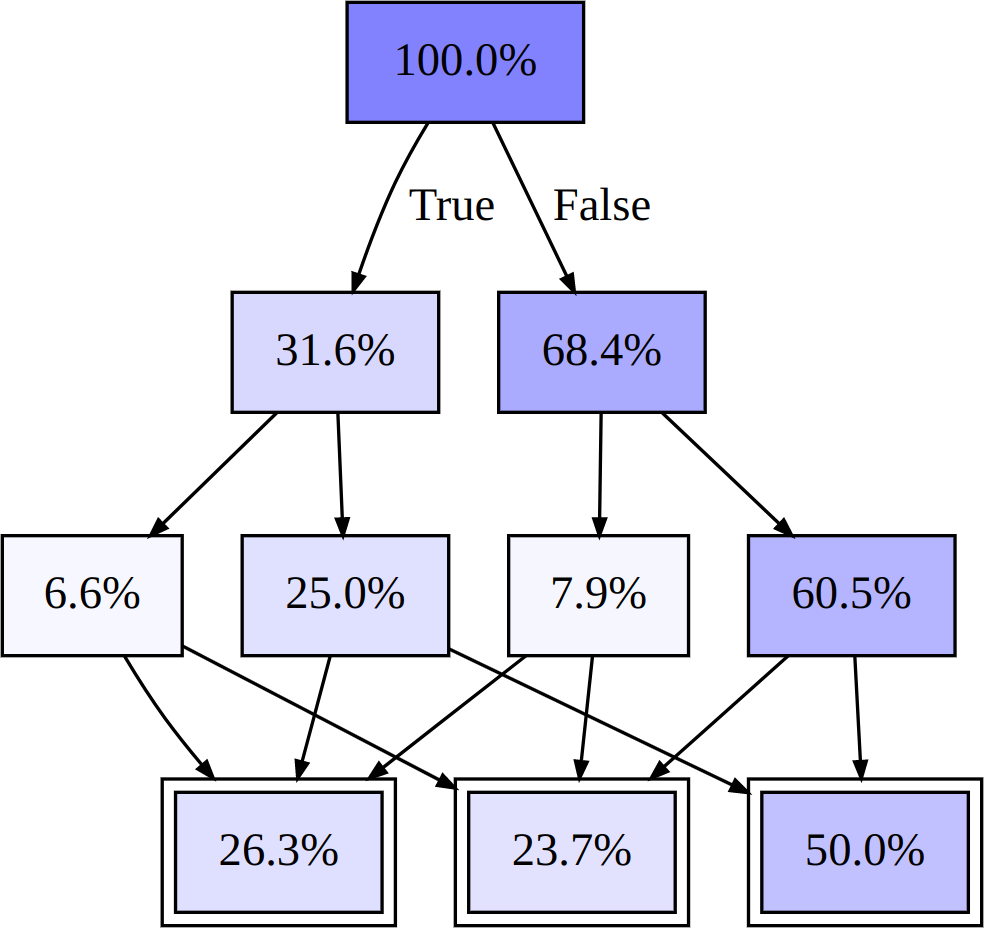}
        \caption{ODT structure}\label{subfig:odtfrag}
    \end{subfigure}
    ~
    \begin{subfigure}{0.4\textwidth}
    \centering
        \includegraphics[scale=0.126]{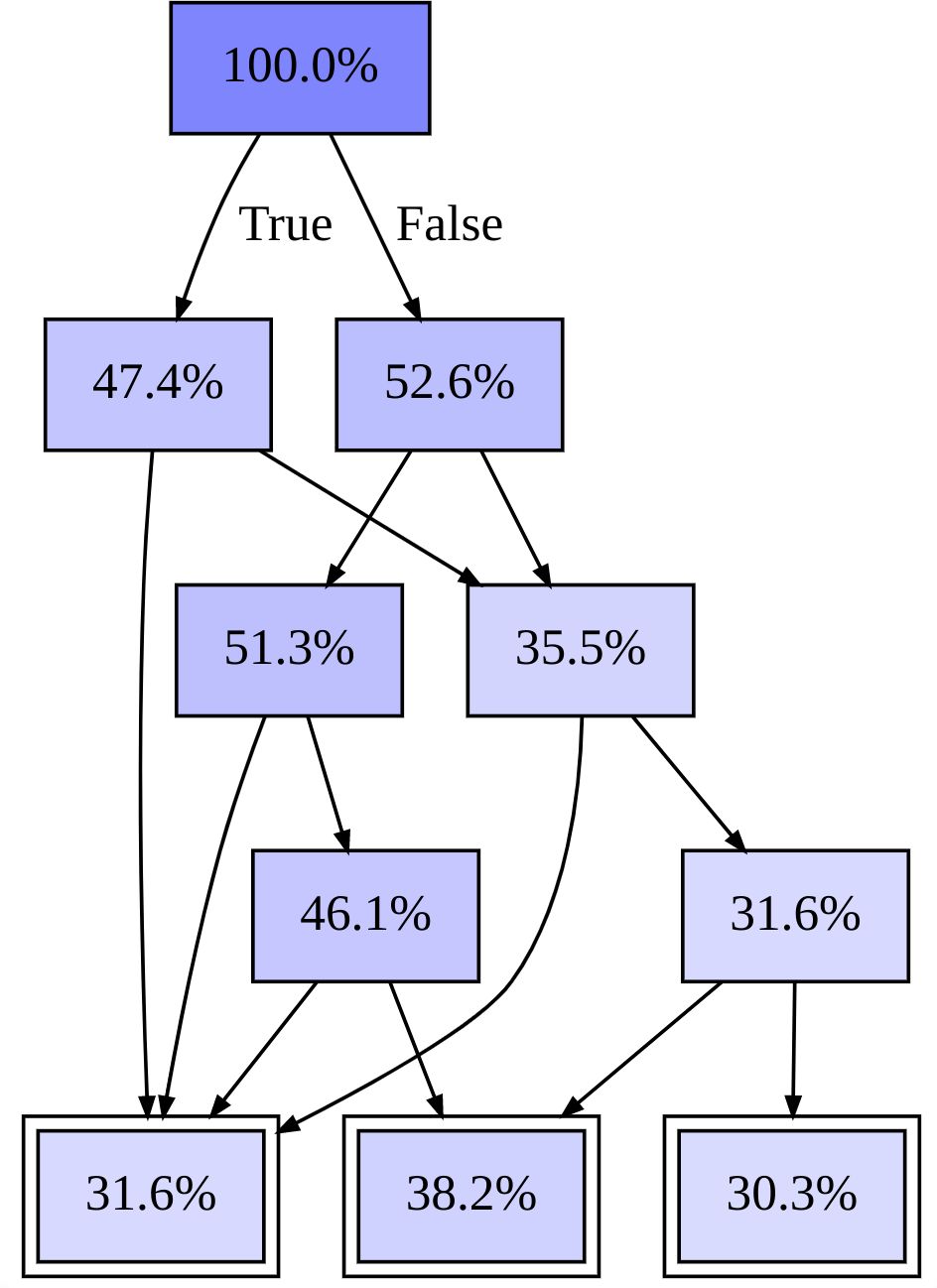}
        \caption{ODD structure}
    \end{subfigure}
    \caption{Fragmentation of an ODT and an ODD trained on the ``teaching-assist. evaluation'' dataset}\label{subfig:oddfrag}
    \label{fig:fragmentation}
\end{figure}

The improved performance of ODDs over ODTs results from the additional freedom in its topology, which permits to express more complex concepts and generally avoids data fragmentation. Figure~\ref{fig:fragmentation} illustrates the resulting different topologies and visualizes the data fragmentation for an ODT and an ODD trained on the ``teaching-assistant evaluation'' dataset. Both models contain seven internal nodes, but the ODD has a much more balanced data fragmentation, i.e., the share of samples processed through each node is more balanced for the ODD.

\paragraph{Stability of ODDs.}
Stability is a desirable characteristic of tree and diagram classifiers, as it preserves effective classification rules in face of new data and enhances model interpretability. Table~\ref{tab:stability} measures the impact on accuracy when enforcing a certain level of stability when training ODDs with our MILP approach. This is achieved by activating Constraints \eqref{eq:stability} and varying the minimum number of samples $S$ required to pass through any active node in the optimal topology. A tradeoff is expected as tighter stability constraints reduce the solution space when constructing the ODD. As observed, more stable ODDs can be obtained at a marginal out-of-sample accuracy loss.

\begin{table}[h]
  \caption{Test accuracy when enforcing minimum flow of samples through active nodes}
  \label{tab:stability}
  \centering
\begin{tabular}{lrrrr}
\toprule
Split & $S=0.05n$ & $0.10n$ & $0.15n$ & $0.20n$ \\
\midrule
Multivariate & 0.828 & 0.827 & 0.823 & 0.823 \\
Univariate & 0.840 & 0.838 & 0.837 & 0.835 \\
\bottomrule
\end{tabular}
\end{table}

\section{Conclusions} \label{sec:conclusion}
We studied the training of optimal decision diagrams from a combinatorial optimization perspective. Specifically, we proposed the first MILP that allows to train optimal decision diagrams for classification. We conducted an extensive numerical study on $54$ benchmark datasets reflecting a wide variety of practical classification tasks. The degrees of freedom of the model permit to find good ODD topologies that exhibit less data fragmentation than ODTs. Concerning out-of-sample accuracy, we showed that ODDs perform favorably when compared to ODTs for the case of multivariate splits.

The flexibility of our mathematical programming approach permits to extend the proposed model to address a variety of important side requirements. Therefore, we believe that it can serve as an important building block for more elaborate models, suited for a variety of application domains. The proposed optimization model finds its limits in very large datasets with several thousands of data points. As future research perspective, we suggest the investigation of heuristic construction techniques that permit to generate decision diagrams for classification independently of dataset size.

\begin{ack}
This research was enabled in part by support provided by Calcul Québec and Compute Canada.
\end{ack}

{
\small
\bibliographystyle{ormsv080-noURLDOI}
\bibliography{references}

\begin{thebibliography}{53}
\expandafter\ifx\csname natexlab\endcsname\relax\def\natexlab#1{#1}\fi
\expandafter\ifx\csname url\endcsname\relax
  \def\url#1{{\tt #1}}\fi
\expandafter\ifx\csname urlprefix\endcsname\relax\def\urlprefix{URL }\fi
\expandafter\ifx\csname urlstyle\endcsname\relax
  \expandafter\ifx\csname doi\endcsname\relax
  \def\doi#1{doi:\discretionary{}{}{}#1}\fi \else
  \expandafter\ifx\csname doi\endcsname\relax
  \def\doi{doi:\discretionary{}{}{}\begingroup \urlstyle{rm}\Url}\fi \fi

\bibitem[{Ab{\'i}o et~al.(2012)Ab{\'i}o, Nieuwenhuis, Oliveras,
  Rodr{\'i}guez-Carbonell, and Mayer-Eichberger}]{Abio2012}
Ab{\'i}o, I., R.~Nieuwenhuis, A.~Oliveras, E.~Rodr{\'i}guez-Carbonell,
  V.~Mayer-Eichberger. 2012.
\newblock A new look at {B}{D}{D}s for pseudo-boolean constraints.
\newblock {\it Journal of Artificial Intelligence Research\/} {\bf 45}
  443--480.

\bibitem[{Aghaei et~al.(2019)Aghaei, Azizi, and Vayanos}]{Aghaei2019}
Aghaei, S., M.J. Azizi, P.~Vayanos. 2019.
\newblock Learning optimal and fair decision trees for non-discriminative
  decision-making.
\newblock {\it Proceedings of the AAAI Conference on Artificial
  Intelligence\/}, vol.~33. 1418--1426.

\bibitem[{Aghaei et~al.(2020)Aghaei, Gomez, and Vayanos}]{Aghaei2020}
Aghaei, S., A.~Gomez, P.~Vayanos. 2020.
\newblock Learning optimal classification trees: Strong max-flow formulations.
\newblock {\it arXiv preprint arXiv:2002.09142\/} .

\bibitem[{Aglin et~al.(2020)Aglin, Nijssen, and Schaus}]{Aglin2020}
Aglin, G., S.~Nijssen, P.~Schaus. 2020.
\newblock {Learning optimal decision trees using caching branch-and-bound
  search}.
\newblock {\it Proceedings of the AAAI Conference on Artificial Intelligence\/}
  {\bf 34}(4) 3146--3153.

\bibitem[{Akers(1978)}]{akers1978}
Akers, Sheldon~B. 1978.
\newblock Binary decision diagrams.
\newblock {\it IEEE Transactions on computers\/} {\bf 27}(06) 509--516.

\bibitem[{Andersen et~al.(2007)Andersen, Hadzic, Hooker, and
  Tiedemann}]{Andersen2007}
Andersen, H.R., T.~Hadzic, J.N. Hooker, P.~Tiedemann. 2007.
\newblock A constraint store based on multivalued decision diagrams.
\newblock {\it International Conference on Principles and Practice of
  Constraint Programming\/}. Springer, 118--132.

\bibitem[{Behle(2007)}]{Behle2007}
Behle, M. 2007.
\newblock {Binary Decision Diagrams and Integer Programming}.
\newblock Ph.D. thesis, Saarland University.

\bibitem[{Belotti et~al.(2016)Belotti, Bonami, Fischetti, Lodi, Monaci,
  Nogales-G{\'{o}}mez, and Salvagnin}]{Belotti2016}
Belotti, P., P.~Bonami, M.~Fischetti, A.~Lodi, M.~Monaci,
  A.~Nogales-G{\'{o}}mez, D.~Salvagnin. 2016.
\newblock {On handling indicator constraints in mixed integer programming}.
\newblock {\it Computational Optimization and Applications\/} {\bf 65}(3)
  545--566.

\bibitem[{Bergman et~al.(2016)Bergman, Cire, {Van Hoeve}, and
  Hooker}]{Bergman2016}
Bergman, D., A.A. Cire, W.-J. {Van Hoeve}, J.~Hooker. 2016.
\newblock {\it {Decision diagrams for optimization}\/}.
\newblock Springer International Publishing.

\bibitem[{Bertsimas and Dunn(2017)}]{Bertsimas2017}
Bertsimas, Dimitris, Jack Dunn. 2017.
\newblock {Optimal classification trees}.
\newblock {\it Machine Learning\/} {\bf 106}(7) 1039--1082.

\bibitem[{Biddle(2006)}]{Biddle2006}
Biddle, D. 2006.
\newblock {\it Adverse impact and test validation: A practitioner's guide to
  valid and defensible employment testing\/}.
\newblock Gower Publishing, Ltd.

\bibitem[{Bixby(2012)}]{Bixby2012}
Bixby, R.E. 2012.
\newblock {A brief history of linear and mixed-integer programming
  computation}.
\newblock {\it Documenta Mathematica\/}  107--121.

\bibitem[{Blanquero et~al.(2020)Blanquero, Carrizosa, Molero-R{\'{i}}o, and
  {Romero Morales}}]{Blanquero2020}
Blanquero, R., E.~Carrizosa, C.~Molero-R{\'{i}}o, D.~{Romero Morales}. 2020.
\newblock {Sparsity in optimal randomized classification trees}.
\newblock {\it European Journal of Operational Research\/} {\bf 284} 255--272.

\bibitem[{Bollig and Wegener(1996)}]{Bollig1996}
Bollig, B., I.~Wegener. 1996.
\newblock {Improving the variable ordering of OBDDs is NP-complete}.
\newblock {\it IEEE Transactions on Computers\/} {\bf 45}(9) 993--1002.

\bibitem[{Breiman et~al.(1984)Breiman, Friedman, Olshen, and
  Stone}]{Breiman1984a}
Breiman, L., J.H. Friedman, R.A. Olshen, C.J. Stone. 1984.
\newblock {\it {Classification and regression trees}\/}.

\bibitem[{Breslow and Aha(1997)}]{Breslow1997}
Breslow, L.A., D.W. Aha. 1997.
\newblock {Simplifying decision trees: A survey}.
\newblock {\it Knowledge Engineering Review\/} {\bf 12}(1) 1--40.

\bibitem[{Bryant(1986)}]{Bryant1986}
Bryant, R.E. 1986.
\newblock {Graph-based algorithms for boolean function manipulation}.
\newblock {\it IEEE Transactions on Computers\/} {\bf C-35}(8) 677--691.

\bibitem[{Bryant(1992)}]{Bryant1992}
Bryant, R.E. 1992.
\newblock {Symbolic boolean manipulation with ordered binary-decision
  diagrams}.
\newblock {\it ACM Computing Surveys\/} {\bf 24}(3) 293--318.

\bibitem[{Cabodi et~al.(2021)Cabodi, Camurati, Ignatiev, Marques-Silva, Palena,
  and Pasini}]{cabodi2021}
Cabodi, Gianpiero, Paolo~E Camurati, Alexey Ignatiev, Joao Marques-Silva, Marco
  Palena, Paolo Pasini. 2021.
\newblock Optimizing binary decision diagrams for interpretable machine
  learning classification.
\newblock {\it 2021 Design, Automation \& Test in Europe Conference \&
  Exhibition (DATE)\/}. IEEE, 1122--1125.

\bibitem[{Carrizosa et~al.(2021)Carrizosa, Molero-R{\'\i}o, and
  Morales}]{Carrizosa2020}
Carrizosa, E., C.~Molero-R{\'\i}o, D.R. Morales. 2021.
\newblock Mathematical optimization in classification and regression trees.
\newblock {\it Top\/} {\bf 29}(1) 5--33.

\bibitem[{Castro et~al.(2019)Castro, Piacentini, Cire, and Beck}]{Castro2019}
Castro, M.P., C.~Piacentini, A.A. Cire, J.C. Beck. 2019.
\newblock Relaxed bdds: An admissible heuristic for delete-free planning based
  on a discrete relaxation.
\newblock {\it Proceedings of the International Conference on Automated
  Planning and Scheduling\/}, vol.~29. 77--85.

\bibitem[{Chorowski and Zurada(2011)}]{Chorowski2011}
Chorowski, J., J.M. Zurada. 2011.
\newblock Extracting rules from neural networks as decision diagrams.
\newblock {\it IEEE Transactions on Neural Networks\/} {\bf 22}(12) 2435--2446.

\bibitem[{Choudhary et~al.(2020)Choudhary, Mishra, Goswami, and
  Sarangapani}]{Choudhary2020}
Choudhary, T., V.~Mishra, A.~Goswami, J.~Sarangapani. 2020.
\newblock A comprehensive survey on model compression and acceleration.
\newblock {\it Artificial Intelligence Review\/}  1--43.

\bibitem[{Demirovi{\'c} et~al.(2020)Demirovi{\'c}, Lukina, Hebrard, Chan,
  Bailey, Leckie, Ramamohanarao, and Stuckey}]{Demirovic2020}
Demirovi{\'c}, E., A.~Lukina, E.~Hebrard, J.~Chan, J.~Bailey, C.~Leckie,
  K.~Ramamohanarao, P.J. Stuckey. 2020.
\newblock Murtree: Optimal classification trees via dynamic programming and
  search.
\newblock {\it arXiv preprint arXiv:2007.12652\/} .

\bibitem[{Dua and Graff(2017)}]{Dua2019}
Dua, D., C.~Graff. 2017.
\newblock {UCI} machine learning repository.

\bibitem[{Firat et~al.(2020)Firat, Crognier, Gabor, Hurkens, and
  Zhang}]{Firat2020}
Firat, M., G.~Crognier, A.F. Gabor, C.A.J. Hurkens, Y.~Zhang. 2020.
\newblock {Column generation based heuristic for learning classification
  trees}.
\newblock {\it Computers and Operations Research\/} {\bf 116} 104866.

\bibitem[{Gossen and Steffen(2019)}]{Gossen2019}
Gossen, F., B.~Steffen. 2019.
\newblock Large random forests: Optimisation for rapid evaluation.
\newblock {\it arXiv preprint arXiv:1912.10934\/} .

\bibitem[{Hu et~al.(2022)Hu, Huguet, and Siala}]{hu2022optimizing}
Hu, Hao, Marie-Jos{\'e} Huguet, Mohamed Siala. 2022.
\newblock Optimizing binary decision diagrams with maxsat for classification .

\bibitem[{Hyafil and Rivest(1976)}]{Hyafil1976}
Hyafil, L., R.L. Rivest. 1976.
\newblock {Constructing optimal decision trees is NP-complete}.
\newblock {\it Information Processing Letters\/} {\bf 5}(1) 1--3.

\bibitem[{Ignatov and Ignatov(2018)}]{Ignatov2018}
Ignatov, D., A.~Ignatov. 2018.
\newblock {Decision stream: Cultivating deep decision trees}.
\newblock {\it Proceedings - International Conference on Tools with Artificial
  Intelligence, ICTAI\/}  905--912.

\bibitem[{Kohavi(1994)}]{Kohavi1994}
Kohavi, R. 1994.
\newblock {Bottom-up induction of oblivious read-once decision graphs:
  strengths and limitations}.
\newblock {\it Proceedings of AAAI\/}  613--618.

\bibitem[{Kohavi and Li(1995)}]{Kohavi1995}
Kohavi, R., C.-H. Li. 1995.
\newblock {Oblivious decision trees, graphs, and top-down pruning}.
\newblock {\it Proceedings of IJCAI\/}. 1071--1079.

\bibitem[{Kumar et~al.(2017)Kumar, Goyal, and Varma}]{Kumar2017}
Kumar, A., S.~Goyal, M.~Varma. 2017.
\newblock {Resource-efficient machine learning in 2 KB RAM for the Internet of
  Things}.
\newblock {\it 34th International Conference on Machine Learning, ICML 2017\/}
  {\bf 4} 3062--3071.

\bibitem[{Lai et~al.(2013)Lai, Liu, and Wang}]{Lai2013}
Lai, Y., D.~Liu, S.~Wang. 2013.
\newblock Reduced ordered binary decision diagram with implied literals: A new
  knowledge compilation approach.
\newblock {\it Knowledge and Information Systems\/} {\bf 35} 665--712.

\bibitem[{Lange and Swoboda(2020)}]{Lange2020}
Lange, Jan-Hendrik, Paul Swoboda. 2020.
\newblock Efficient message passing for 0-1 {ILPs} with binary decision
  diagrams .

\bibitem[{Lee(1959)}]{Lee1959}
Lee, C.~Y. 1959.
\newblock {Representation of Switching Circuits by Binary‐Decision Programs}.
\newblock {\it Bell System Technical Journal\/} {\bf 38}(4) 985--999.

\bibitem[{Mehrabi et~al.(2019)Mehrabi, Morstatter, Saxena, Lerman, and
  Galstyan}]{Mehrabi2019}
Mehrabi, N., F.~Morstatter, N.~Saxena, K.~Lerman, A.~Galstyan. 2019.
\newblock A survey on bias and fairness in machine learning.
\newblock {\it arXiv preprint arXiv:1908.09635\/} .

\bibitem[{Meisel and Michalopoulos(1973)}]{Meisel1973}
Meisel, W.S., D.A. Michalopoulos. 1973.
\newblock {A partitioning algorithm with application in pattern classification
  and the optimization of decision trees}.
\newblock {\it IEEE Transactions on Computers\/} {\bf C-22}(1) 93--103.

\bibitem[{Mues et~al.(2004)Mues, Baesens, Files, and Vanthienen}]{Mues2004}
Mues, Christophe, Bart Baesens, Craig~M. Files, Jan Vanthienen. 2004.
\newblock {Decision diagrams in machine learning: An empirical study on
  real-life credit-risk data}.
\newblock {\it Expert Systems with Applications\/} {\bf 27}(2) 257--264.

\bibitem[{Narodytska et~al.(2018)Narodytska, Ignatiev, Pereira, and
  Marques-Silva}]{Narodytska2018}
Narodytska, N., A.~Ignatiev, F.~Pereira, J.~Marques-Silva. 2018.
\newblock {Learning optimal decision trees with SAT}.
\newblock {\it Proceedings of the Twenty-Seventh International Joint Conference
  on Artificial Intelligence, IJCAI-18\/}. 1362--1368.

\bibitem[{Oliveira and Sangiovanni-Vincentelli(1996)}]{Oliveira1996}
Oliveira, A.L., A.~Sangiovanni-Vincentelli. 1996.
\newblock {Using the minimum description length principle to infer reduced
  ordered decision graphs}.
\newblock {\it Machine Learning\/} {\bf 25}(1) 23--50.

\bibitem[{Oliver(1993)}]{Oliver1993}
Oliver, J. 1993.
\newblock {Decision graphs -- An extension of decision trees}.
\newblock {\it Proceedings of the 4th international workshop on artificial
  intelligence and statistics (AISTATS)\/}. 343--–350.

\bibitem[{Payne and Meisel(1977)}]{Payne1977}
Payne, H.J., W.S. Meisel. 1977.
\newblock An algorithm for constructing optimal binary decision trees.
\newblock {\it IEEE Computer Architecture Letters\/} {\bf 26}(09) 905--916.

\bibitem[{Perez and R{\'e}gin(2015)}]{Perez2015}
Perez, G., J.-C. R{\'e}gin. 2015.
\newblock Efficient operations on mdds for building constraint programming
  models.
\newblock {\it IJCAI 2015\/}.

\bibitem[{Platt et~al.(2000)Platt, Cristianini, and Shawe-Taylor}]{Platt2000}
Platt, J.C., N.~Cristianini, J.~Shawe-Taylor. 2000.
\newblock {Large margin DAGs for multiclass classification}.
\newblock {\it Advances in Neural Information Processing Systems\/}  547--553.

\bibitem[{Sanner et~al.(2010)Sanner, Uther, and Delgado}]{Sanner2010}
Sanner, S., W.T.B. Uther, K.V. Delgado. 2010.
\newblock Approximate dynamic programming with affine adds.
\newblock {\it AAMAS\/}. 1349--1356.

\bibitem[{Serra(2020)}]{Serra2020}
Serra, T. 2020.
\newblock Enumerative branching with less repetition.
\newblock {\it International Conference on Integration of Constraint
  Programming, Artificial Intelligence, and Operations Research\/}. Springer,
  399--416.

\bibitem[{Shotton et~al.(2013)Shotton, Nowozin, Sharp, Winn, Kohli, and
  Criminisi}]{Shotton2013}
Shotton, J., S.~Nowozin, T.~Sharp, J.~Winn, P.~Kohli, A.~Criminisi. 2013.
\newblock {Decision jungles: Compact and rich models for classification}.
\newblock {\it Advances in Neural Information Processing Systems\/}  1--9.

\bibitem[{Verhaeghe et~al.(2018)Verhaeghe, Lecoutre, and
  Schaus}]{Verhaeghe2018}
Verhaeghe, H., C.~Lecoutre, P.~Schaus. 2018.
\newblock {Compact-MDD: Efficiently filtering (s)MDD constraints with
  reversible sparse bit-sets}.
\newblock {\it IJCAI\/}. 1383--1389.

\bibitem[{Verhaeghe et~al.(2020)Verhaeghe, Nijssen, Pesant, Quimper, and
  Schaus}]{Verhaeghe2020}
Verhaeghe, H., S.~Nijssen, G.~Pesant, C.-G. Quimper, P.~Schaus. 2020.
\newblock {Learning optimal decision trees using constraint programming}.
\newblock {\it Proceedings of the Twenty-Ninth International Joint Conference
  on Artificial Intelligence, IJCAI-20\/}. 4765--4769.

\bibitem[{Vidal and Schiffer(2020)}]{Vidal2020a}
Vidal, T., M.~Schiffer. 2020.
\newblock {Born-Again Tree Ensembles}.
\newblock Hal~Daum{\'{e}} III, Aarti Singh, eds., {\it Proceedings of the 37th
  International Conference on Machine Learning\/}, {\it Proceedings of Machine
  Learning Research\/}, vol. 119. PMLR, Virtual, 9743--9753.

\bibitem[{Ye and Xie(2020)}]{Ye2020}
Ye, Q., W.~Xie. 2020.
\newblock Unbiased subdata selection for fair classification: A unified
  framework and scalable algorithms.
\newblock {\it arXiv preprint arXiv:2012.12356\/} .

\bibitem[{Zafar et~al.(2017)Zafar, Valera, Rodriguez, and Gummadi}]{Zafar2017}
Zafar, M.B., I.~Valera, M.G. Rodriguez, K.P. Gummadi. 2017.
\newblock {Fairness constraints: Mechanisms for fair classification}.
\newblock {\it Proceedings of the 20th International Conference on Artificial
  Intelligence and Statistics, AISTATS 2017\/} {\bf 54}.

\end{thebibliography}
}


\appendix

\renewcommand{\smash}[1]{#1}


\section{Proof of Theorems}
\paragraph{Theorem~1.}
Formulation (\ref{start:formulation}--\ref{end:formulation}) produces solutions in which all variables $\mathbf{w}$ and $\mathbf{z}$ take binary values, leading to a feasible and optimal decision diagram.

\paragraph{Proof of Theorem~1.}
The proof goes by induction. More precisely, we first demonstrate the following statement:

\noindent
\emph{For each sample $i$ and each layer \smash{$l \in \{0,\dots,D-1\}$}, at most a single \smash{$u \in \cVI_l$} receives a unit of flow from this sample, i.e., in such a way that \smash{$w^+_{iu} + w^-_{iu} = 1$}, while the other nodes receive no flow, i.e., such that \smash{$w^+_{iu} + w^-_{iu} = 0$}.
}

We start with the base case of layer $0$ containing a single node, the root node, which immediately satisfies $w^+_{iu} + w^-_{iu} = 1$ due to Equation~\eqref{start:formulation}. The induction step then goes as follows. Suppose that there exists a single node $\bar{u}$ at layer $l$ such that $w^+_{i\bar{u}} + w^-_{i\bar{u}} = 1$, as a consequence:
\begin{itemize}[leftmargin=*]
\item If $\lambda_{il} = 1$, then \smash{$\sum_{u \in \cVI_{l}} w^-_{iu} \leq 0$} due to Equation~\eqref{eq:my4} and therefore \smash{$w^-_{i\bar{u}} = 0$}. Since \smash{$w^+_{i\bar{u}} + w^-_{i\bar{u}} = 1$}, we have \smash{$w^+_{i\bar{u}} = 1$}. Next, since variables~\smash{$\mathbf{y}^+$} are binary and due to Equation~\eqref{myeq6}, at most one design variable \smash{$y_{\bar{u}\bar{v}}^+$} takes value $1$ for one specific $\bar{v}$.
Along with Equation~\eqref{myeq9}, this implies \smash{$z_{i\bar{u}v}^+ = 0$} for all $v \neq \bar{v}$. Finally, due to Equation~\eqref{myeq3}, we have \smash{$w^+_{i\bar{u}} = 1 = \sum_{v \in \delta^+(\bar{u})} z^+_{i\bar{u}v}$} and therefore \smash{$z^+_{i\bar{u}\bar{v}}$ = 1}. This means that $\bar{u} \in \cVI_l$ sends exactly one unit of flow to \smash{$\bar{v} \in \cVI_{l+1} \cup \cVC$}. Next, observe that \smash{$w^+_{iu} + w^-_{iu} = 0$} for \smash{$u \neq \bar{u}$} implies that \smash{$z^+_{i u v} = 0$} and \smash{$z^-_{i u v} = 0$} for all \smash{$v \in \cVI_{l+1} \cup \cVC$}, and therefore the other nodes from layer $l$ do not send any flow.

\item If $\lambda_{il} = 0$, then \smash{$\sum_{u \in \cVI_{l}} w^+_{iu} \leq 0$} due to Equation~\eqref{eq:my5}
and therefore \smash{$w^+_{i\bar{u}} = 0$}. Since \smash{$w^+_{i\bar{u}} + w^-_{i\bar{u}} = 1$}, we have \smash{$w^-_{i\bar{u}} = 1$}. The rest of the reasoning is identical to the previous case, using the negative-side instead of the positive-side flow along with the corresponding equations.
\end{itemize}

Overall, this means that $\bar{v}$ is the only possible node in layer $\cVI_{l+1}$ with one unit of flow for sample $i$, therefore completing the induction step.

Observe that variables \smash{$w^+_{iu}$} and \smash{$w^-_{iu}$} take binary values in both cases discussed in the induction step. Moreover, we have \smash{$z^+_{iuv} = 1$} if and only if \smash{$w^+_{iu} = 1$} and \smash{$y_{uv} = 1$}, otherwise \smash{$z^+_{iuv} = 0$}. As such, the integrality of the $\mathbf{z}$ variables is implied by the integrality of the $\mathbf{w}$ variables.
As a consequence, each sample $i$ follows a single path within the decision diagram down to a terminal node determined by the $\mathbf{w}$ and $\mathbf{z}$ variables. Finally, the branch choice at each node (negative-side or positive-side path) represented by the $\mathbf{w}$ and $\bf{\lambda}$ variables is directly determined by the position of the sample relative to the hyperplane due to Equations~\eqref{eq:logic1} and \eqref{eq:logic2}. Overall, this ensures that the MILP represents the space of all possible valid models and, in the meanwhile, correctly evaluates the trajectory of the samples within them to obtain the correct accuracy. An optimal solution of this MILP using state-of-the-art branch-and-cut solvers therefore permits to train the model to global optimality.

\paragraph{Theorem~2.}
Our formulation includes $\cO(n \log|\cV|)$ binary decision variables when applied to decision-tree skeletons.

\paragraph{Proof of Theorem~2.} 
Our model contains only two families of decision variables which must be defined as integer (and require possible branching in a branch-and-cut based algorithm): the binary design variables $\mathbf{y}$ and the branch variables $\bf{\lambda}$. The integrality of the rest of the variables directly derives from the integrality of the former ones, as seen in Theorem~1.

A decision-tree skeleton is such that $|\cVI_l| = 2^l$ for each layer $l \in \{0,\dots,D-1\}$. For such a case, the number of $\mathbf{y}$ variables of our model is $\smash{2 \sum_{l=0}^{D-1} (2^l (2^{l+1} + |\cC|))} = \smash{\cO(4^D + 2^D |\cC|)} = \smash{\cO(|\cV|(|\cV|+|\cC|)}$, whereas the number of $\bf{\lambda}$ variables is in $\cO(n D) = \cO(n \log |V|)$. We can reasonably assume that the number of samples is largely superior to $|\cV|$ and $|\cC|$, such that $nD \gg |\cV|(|\cV|+|\cC|)$. As such, the overall number of integer variables is $\cO(n \log |V|)$.

In comparison, the formulation of~\citep{Bertsimas2017} creates a complete layer containing $2^D$ terminal nodes, and associates to each terminal node and sample a binary variable to determine if the sample reaches this node. This leads to $\cO(n|\cV|)$ binary decision variables defining the samples trajectories. In a similar fashion, there exists other integer variables in the model (e.g., variables defining the class of each terminal node) which are less numerous under the assumption that $n \gg |\cC|$ and $n \gg |\cV|$.

\section{Detailed Computational Results}

In this section, we provide more details about the numerical study. In particular, we present the list of datasets used in the experiments, additional results regarding computational performance and hyperparameters tuning, and an accuracy comparison of ODTs and ODDs per dataset.

\subsection{Dataset list}
\label{sec:datasets}

Table \ref{tab:datasets} lists the 54 datasets on which we conducted the numerical experiments. These datasets are the same as considered by \citet{Bertsimas2017}. The 18 datasets indicated with bold font satisfy the following conditions:
\begin{enumerate}
\item Datasets where the MILP model can consistently improve upon the solution found in Step 1 of the training strategy. Specifically, we select datasets where the model produced a better solution in at least 125 out of the 250 model runs, considering all combinations of split type (univariate and multivariate), random seed, skeleton, and value of~$\alpha$.
\item Datasets where the best topology found after Step 2 of the training strategy contains at least~$|\cVC|$ internal nodes.
\end{enumerate}

Finding ODTs and ODDs in the datasets that do not satisfy condition 1 is a significant challenge, considering also the short computational budget allowed (600 seconds of CPU time). Furthermore, datasets that do not satisfy condition 2 are less interesting because the training data can be separated in many cases with a single split at the root node. Hence, the 18 selected datasets allow a more faithful comparison of the accuracy of ODTs and ODDs, as shown in Figures \ref{fig:boxPerformance} and \ref{fig:performanceSort}. We also provide detailed accuracy comparison over all 54 datasets in Section \ref{sec:accuracy}.

\begin{table}
  \caption{The 54 datasets used in the computational experiments}
  \label{tab:datasets}
  \centering
\begin{tabular}{lrrrllrrr}
\toprule
Dataset & $n$ & $d$ & \small{$|\cVC|$} &  & Dataset & $n$ & $d$ & \small{$|\cVC|$} \\
\midrule
acute-inflam-nephritis & 120 & 6 & 2 &  & iris & 150 & 4 & 3 \\
acute-inflam-urinary & 120 & 6 & 2 &  & mammographic-mass & 830 & 10 & 2 \\
balance-scale & 625 & 4 & 3 &  & monks1 & 556 & 11 & 2 \\
\textbf{banknote-auth} & 1372 & 4 & 2 &  & \textbf{monks2} & 601 & 11 & 2 \\
\textbf{blood-transfusion} & 748 & 4 & 2 &  & monks3 & 554 & 11 & 2 \\
breast-cancer-diag & 569 & 30 & 2 &  & optical-recognition & 5620 & 64 & 10 \\
\textbf{breast-cancer-prog} & 194 & 33 & 2 &  & ozone-eighthr & 1847 & 72 & 2 \\
breast-cancer-wisconsin & 699 & 9 & 2 &  & ozone-onehr & 1848 & 72 & 2 \\
car-evaluation & 1728 & 15 & 4 &  & \textbf{parkinsons} & 195 & 22 & 2 \\
chess-kr-vs-kp & 3196 & 37 & 2 &  & \textbf{pima-indians-diab} & 768 & 8 & 2 \\
\textbf{climate-simul-crashes} & 540 & 18 & 2 &  & \textbf{planning-relax} & 182 & 12 & 2 \\
congressional-voting & 435 & 16 & 2 &  & \textbf{qsar-biodegradation} & 1055 & 41 & 2 \\
\textbf{connect-mines-rocks} & 208 & 60 & 2 &  & seeds & 210 & 7 & 3 \\
connect-vowel & 990 & 10 & 11 &  & \textbf{seismic-bumps} & 2584 & 20 & 2 \\
contraceptive-method & 1473 & 11 & 3 &  & soybean-small & 47 & 35 & 4 \\
\textbf{credit-approval} & 653 & 37 & 2 &  & spambase & 4601 & 57 & 2 \\
\textbf{cylinder-bands} & 277 & 484 & 2 &  & \textbf{spect-heart} & 267 & 22 & 2 \\
dermatology & 366 & 34 & 6 &  & \textbf{spectf-heart} & 267 & 44 & 2 \\
echocardiogram & 61 & 9 & 2 &  & statlog-german-credit & 1000 & 48 & 2 \\
fertility-diagnosis & 100 & 12 & 2 &  & statlog-landsat-sat & 6435 & 36 & 6 \\
habermans-survival & 306 & 3 & 2 &  & teaching-assist-eval & 151 & 52 & 3 \\
hayes-roth & 160 & 4 & 3 &  & \textbf{thoracic-surgery} & 470 & 24 & 2 \\
heart-disease-cleveland & 297 & 18 & 5 &  & thyroid-ann & 3772 & 21 & 3 \\
hepatitis & 80 & 19 & 2 &  & thyroid-new & 215 & 5 & 3 \\
image-segmentation & 2310 & 18 & 7 &  & \textbf{tic-tac-toe} & 958 & 18 & 2 \\
indian-liver-patient & 579 & 10 & 2 &  & wall-following-robot-2 & 5456 & 2 & 4 \\
\textbf{ionosphere} & 351 & 33 & 2 &  & wine & 178 & 13 & 3 \\
\bottomrule
\end{tabular}
\end{table}

\subsection{Computational performance}
\begin{figure}[h]
	\begin{center}
		\begin{minipage}[t]{0.495\textwidth}
			\centering
			\includegraphics[scale=0.6925,trim={0 0 0 0}]{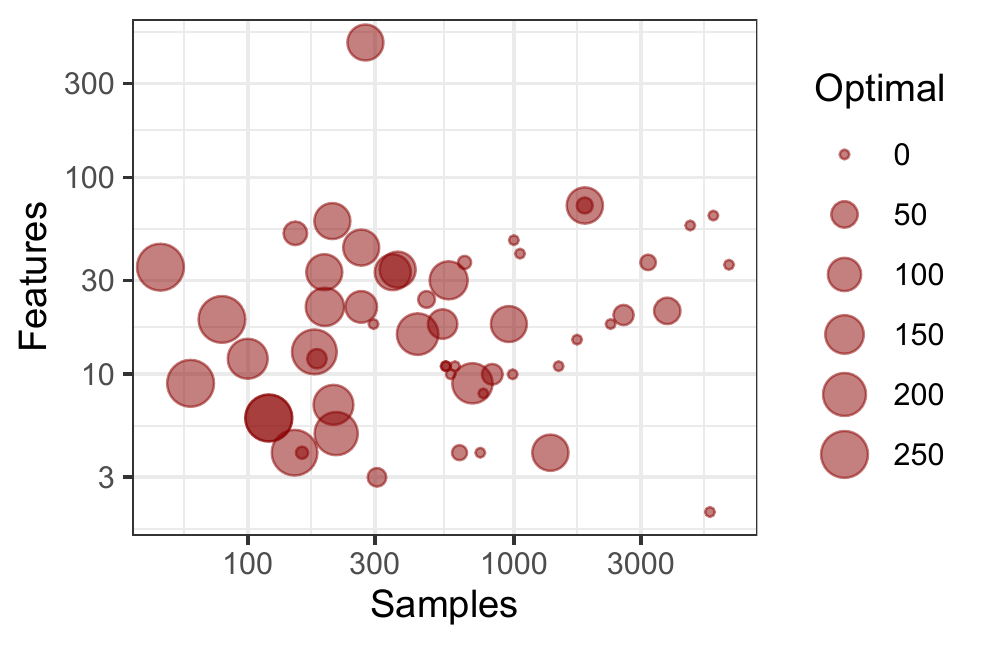}
            \caption{Optimal topologies}\label{fig:optimal}
		\end{minipage}%
		\hfill
		\begin{minipage}[t]{0.495\textwidth}
			\centering
    	    \includegraphics[scale=0.6925,trim={0 0 0 0}]{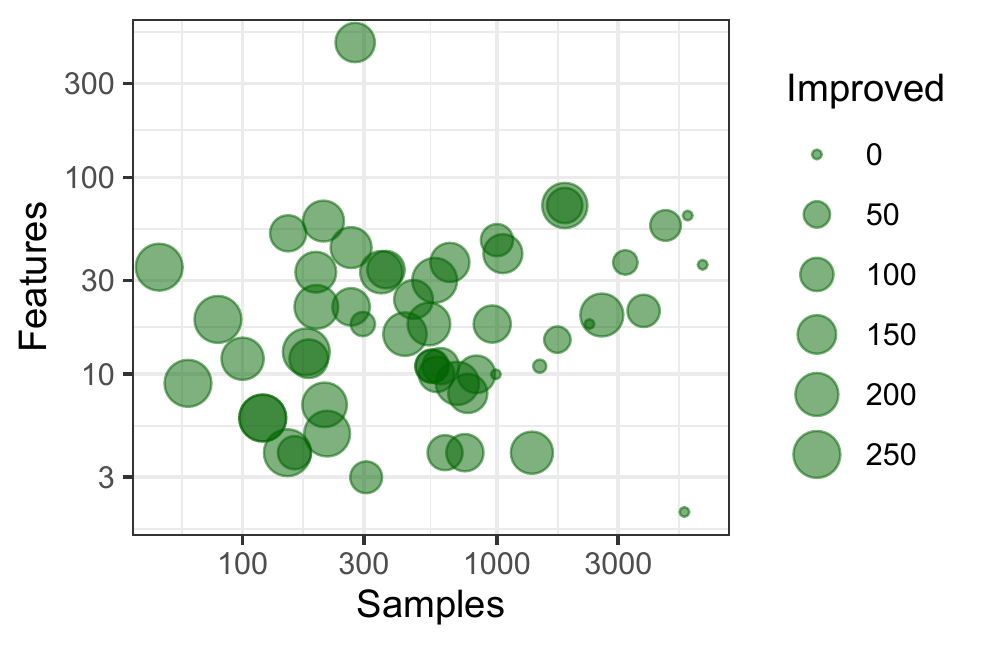}
            \caption{Improved topologies}\label{fig:better}
		\end{minipage}
	\end{center}
\end{figure}

The proposed MILP for training ODDs is effective in either finding optimal topologies or improving the topology found in Step 1 of the training strategy. Figure \ref{fig:optimal} illustrates the number of model runs that resulted in a proven optimal topology as a function of the number of samples and features of each dataset. Note that the MILP is solved 250 times for each dataset (two split types, five alpha values, five topologies -- including the tree topology -- and five seeds). Similarly, Figure \ref{fig:better} shows the number of datasets where the MILP improves upon the solution found in Step 1. As observed, the difficulty of the training problem generally increases with the number of samples. Most distinctly, the method is able to find optimal or improved topologies in high-dimensional datasets with tens or hundreds of features.

\subsection{Hyperparameters tuning}
The hyperparameter $\alpha$ controls overfitting by penalizing the activation of nodes and, consequently, complex topologies. Whereas ODTs are regularized by tuning only $\alpha$, ODDs are regularized by tuning $\alpha$ and also selecting the best skeleton for each dataset and split type, via the same cross-validation procedure.

Table \ref{tab:tuning} shows, out of the 108 combinations of datasets and split type, how many are regularized by each combination of $\alpha$ and skeleton. As seen, a lower $\alpha$ value (which leads to richer topologies) is preferable for both ODTs and ODDs. Furthermore, each considered skeleton for ODDs is selected in at least 21\% of all runs, which shows that, for a given number of internal nodes, the best configuration depends largely on the target dataset.

\begin{table}[h]
  \caption{Hyperparameters tuning: number of datasets $\times$ split type per skeleton and value of $\alpha$}
  \label{tab:tuning}
  \centering
\begin{tabular}{lcrrrrrr}
\toprule
Skeleton & ODD/ODT & $\alpha=0.01$ & 0.1 & 0.2 & 0.5 & 1.0 & Total \\
\midrule
(1-2-4-8) & ODD & 9 & 8 & 8 & 4 & 4 & 33 \\
(1-2-4-4-4) & ODD & 14 & 6 & 3 & 2 & 2 & 27 \\
(1-2-3-3-3-3) & ODD & 8 & 6 & 2 & 4 & 3 & 23 \\
(1-2-2-2-2-2-2-2) & ODD & 7 & 7 & 5 & 3 & 3 & 25 \\
\midrule
(1-2-4-8) & ODT & 37 & 21 & 16 & 14 & 20 & 108 \\
\midrule
Total &  & 75 & 48 & 34 & 27 & 32 & 216 \\
\bottomrule
\end{tabular}
\end{table}

\subsection{Accuracy results}
\label{sec:accuracy}

Table \ref{tab:accuracy} shows the out-of-sample accuracy obtained by ODTs and ODDs in all 54 datasets, after the hyperparameters tuning phase. A paired Wilcoxon signed-rank test indicates that in the case of multivariate splits ODDs are more accurate than ODTs ($p$-value of 0.000895). In the case of univariate splits, both ODDs and ODTs are equally accurate within a significance level of 0.05. A focused analysis on the 18 selected datasets (see Section \ref{sec:datasets}) also reveals a higher accuracy of ODDs in the multivariate splits case ($p$-value of 0.000567): the average accuracy gain in those datasets amounts to 1.9 percentage points.

\begin{table}
  \caption{Out-of-sample accuracy of ODTs and ODDs, per dataset}
  \label{tab:accuracy}
  \centering
\scalebox{0.95}{\begin{tabular}{lrrrrrrrr}
\toprule
 &  & \multicolumn{3}{c}{ODTs} &  & \multicolumn{3}{c}{ODDs} \\
\cmidrule(l){3-5}\cmidrule(l){7-9}
 &  & \multicolumn{2}{c}{Split} & Avg &  & \multicolumn{2}{c}{Split} & Avg \\
\cmidrule(l){3-4}\cmidrule(l){7-8}
Dataset &  & Multi & Uni &  &  & Multi & Uni \\
\midrule
acute-inflammations-nephritis$\quad$ &  & 1.000 & 1.000 & 1.000 & $\quad$ & 1.000 & 1.000 & 1.000 \\
acute-inflammations-urinary &  & 1.000 & 1.000 & 1.000 &  & 0.993 & 1.000 & 0.997 \\
balance-scale &  & 0.904 & 0.690 & 0.797 &  & 0.901 & 0.729 & 0.815 \\
banknote-authentication &  & 0.993 & 0.969 & 0.981 &  & 0.997 & 0.985 & 0.991 \\
blood-transfusion-service &  & 0.774 & 0.773 & 0.774 &  & 0.793 & 0.778 & 0.785 \\
breast-cancer-diagnostic &  & 0.930 & 0.918 & 0.924 &  & 0.941 & 0.917 & 0.929 \\
breast-cancer-prognostic &  & 0.688 & 0.721 & 0.704 &  & 0.717 & 0.708 & 0.713 \\
breast-cancer-wisconsin &  & 0.952 & 0.947 & 0.950 &  & 0.957 & 0.939 & 0.948 \\
car-evaluation &  & 0.880 & 0.751 & 0.816 &  & 0.881 & 0.781 & 0.831 \\
chess-kr-vs-kp &  & 0.975 & 0.939 & 0.957 &  & 0.977 & 0.954 & 0.966 \\
climate-simulation-crashes &  & 0.890 & 0.910 & 0.900 &  & 0.902 & 0.895 & 0.899 \\
congressional-voting &  & 0.932 & 0.956 & 0.944 &  & 0.943 & 0.956 & 0.950 \\
connectionist-mines-vs-rocks &  & 0.715 & 0.731 & 0.723 &  & 0.727 & 0.727 & 0.727 \\
connectionist-vowel &  & 0.381 & 0.381 & 0.381 &  & 0.399 & 0.340 & 0.370 \\
contraceptive-method-choice &  & 0.491 & 0.480 & 0.486 &  & 0.483 & 0.493 & 0.488 \\
credit-approval &  & 0.840 & 0.872 & 0.856 &  & 0.843 & 0.872 & 0.858 \\
cylinder-bands &  & 0.667 & 0.609 & 0.638 &  & 0.655 & 0.661 & 0.658 \\
dermatology &  & 0.913 & 0.891 & 0.902 &  & 0.900 & 0.915 & 0.908 \\
echocardiogram &  & 0.880 & 0.933 & 0.907 &  & 0.933 & 0.947 & 0.940 \\
fertility-diagnosis &  & 0.768 & 0.824 & 0.796 &  & 0.832 & 0.792 & 0.812 \\
habermans-survival &  & 0.758 & 0.734 & 0.746 &  & 0.753 & 0.713 & 0.733 \\
hayes-roth &  & 0.645 & 0.555 & 0.600 &  & 0.655 & 0.670 & 0.663 \\
heart-disease-cleveland &  & 0.549 & 0.546 & 0.547 &  & 0.549 & 0.549 & 0.549 \\
hepatitis &  & 0.800 & 0.890 & 0.845 &  & 0.730 & 0.850 & 0.790 \\
image-segmentation &  & 0.915 & 0.915 & 0.915 &  & 0.910 & 0.913 & 0.912 \\
indian-liver-patient &  & 0.706 & 0.716 & 0.711 &  & 0.703 & 0.716 & 0.710 \\
ionosphere &  & 0.820 & 0.886 & 0.853 &  & 0.852 & 0.902 & 0.877 \\
iris &  & 0.911 & 0.947 & 0.929 &  & 0.932 & 0.958 & 0.945 \\
mammographic-mass &  & 0.817 & 0.822 & 0.820 &  & 0.816 & 0.814 & 0.815 \\
monks1 &  & 0.409 & 0.445 & 0.427 &  & 0.471 & 0.476 & 0.473 \\
monks2 &  & 0.651 & 0.545 & 0.598 &  & 0.699 & 0.533 & 0.616 \\
monks3 &  & 0.403 & 0.461 & 0.432 &  & 0.449 & 0.467 & 0.458 \\
optical-recognition &  & 0.697 & 0.694 & 0.695 &  & 0.740 & 0.698 & 0.719 \\
ozone-eighthr &  & 0.923 & 0.935 & 0.929 &  & 0.926 & 0.935 & 0.931 \\
ozone-onehr &  & 0.938 & 0.970 & 0.954 &  & 0.937 & 0.971 & 0.954 \\
parkinsons &  & 0.788 & 0.837 & 0.812 &  & 0.829 & 0.857 & 0.843 \\
pima-indians-diabetes &  & 0.742 & 0.726 & 0.734 &  & 0.742 & 0.716 & 0.729 \\
planning-relax &  & 0.526 & 0.652 & 0.589 &  & 0.609 & 0.652 & 0.630 \\
qsar-biodegradation &  & 0.833 & 0.811 & 0.822 &  & 0.829 & 0.808 & 0.818 \\
seeds &  & 0.919 & 0.885 & 0.902 &  & 0.908 & 0.877 & 0.892 \\
seismic-bumps &  & 0.928 & 0.929 & 0.928 &  & 0.929 & 0.929 & 0.929 \\
soybean-small &  & 0.967 & 0.983 & 0.975 &  & 0.950 & 0.983 & 0.967 \\
spambase &  & 0.913 & 0.899 & 0.906 &  & 0.918 & 0.905 & 0.912 \\
spect-heart &  & 0.627 & 0.737 & 0.682 &  & 0.660 & 0.737 & 0.699 \\
spectf-heart &  & 0.728 & 0.755 & 0.742 &  & 0.737 & 0.734 & 0.736 \\
statlog-german-credit &  & 0.705 & 0.684 & 0.694 &  & 0.707 & 0.683 & 0.695 \\
statlog-landsat-satellite &  & 0.809 & 0.798 & 0.804 &  & 0.796 & 0.803 & 0.800 \\
teaching-assistant-evaluation &  & 0.521 & 0.400 & 0.461 &  & 0.505 & 0.432 & 0.468 \\
thoracic-surgery &  & 0.797 & 0.831 & 0.814 &  & 0.839 & 0.836 & 0.837 \\
thyroid-ann &  & 0.993 & 0.993 & 0.993 &  & 0.995 & 0.994 & 0.995 \\
thyroid-new &  & 0.937 & 0.944 & 0.941 &  & 0.948 & 0.926 & 0.937 \\
tic-tac-toe &  & 0.963 & 0.807 & 0.885 &  & 0.962 & 0.798 & 0.880 \\
wall-following-robot-2 &  & 1.000 & 1.000 & 1.000 &  & 1.000 & 1.000 & 1.000 \\
wine &  & 0.914 & 0.923 & 0.918 &  & 0.932 & 0.918 & 0.925 \\
\midrule
Average &  & 0.799 & 0.795 & 0.797 &  & 0.809 & 0.799 & 0.804 \\
\bottomrule
\end{tabular}}
\end{table}

\subsection{Data fragmentation}
\begin{figure}[h]
    \centering
    \begin{subfigure}{0.4\textwidth}
    \centering
        \includegraphics[scale=0.18]{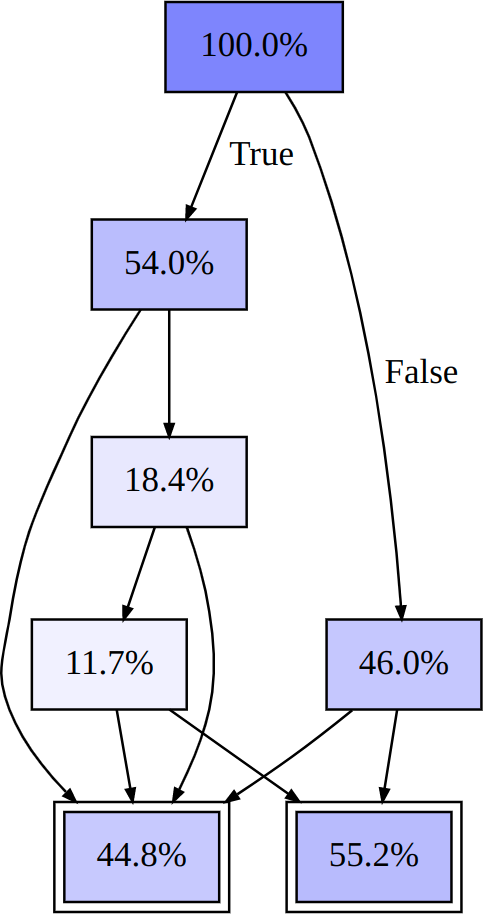}
        \caption{ODT structure}
    \end{subfigure}
    ~
    \begin{subfigure}{0.4\textwidth}
    \centering
        \includegraphics[scale=0.18]{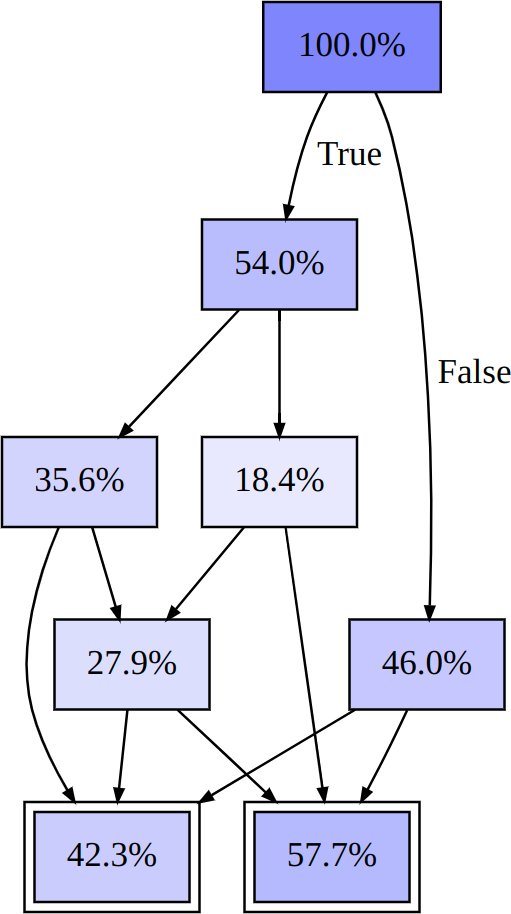}
        \caption{ODD structure}
    \end{subfigure}
    \caption{Fragmentation of an ODT and an ODD trained on the ``credit approval'' dataset}
    \label{fig:frag1}
\end{figure}

\begin{figure}[h]
    \centering
    \begin{subfigure}{0.4\textwidth}
    \centering
        \includegraphics[scale=0.18]{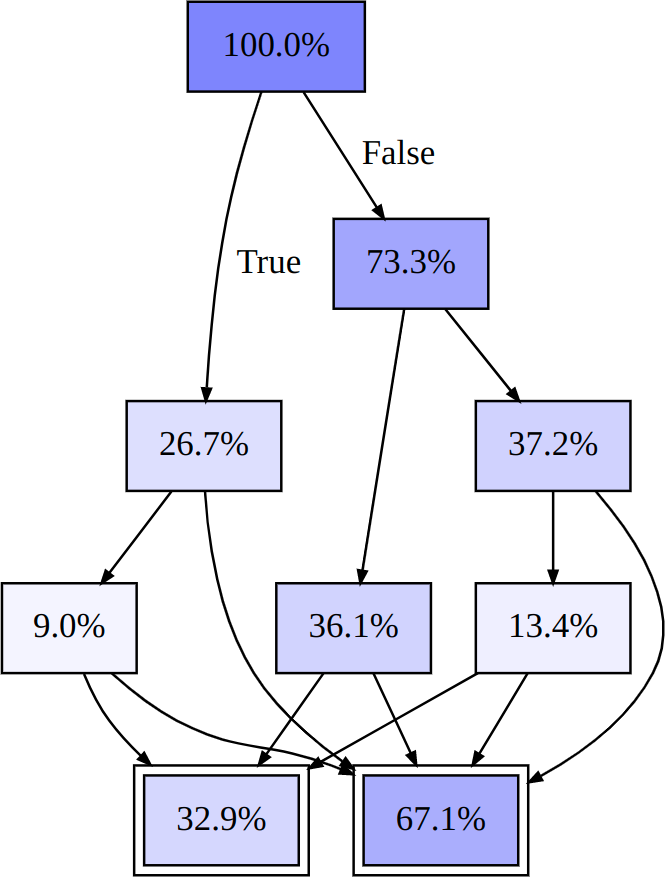}
        \caption{ODT structure}
    \end{subfigure}
    ~
    \begin{subfigure}{0.4\textwidth}
    \centering
        \includegraphics[scale=0.18]{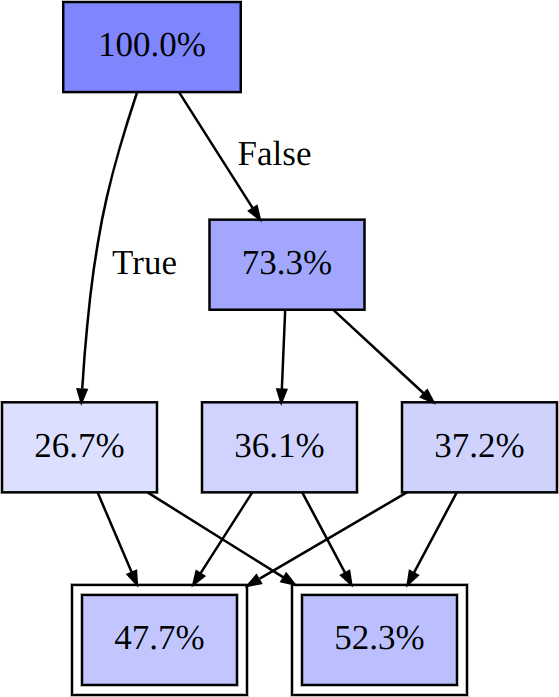}
        \caption{ODD structure}
    \end{subfigure}
    \caption{Fragmentation of an ODT and an ODD trained on the ``monks3'' dataset}
    \label{fig:frag2}
\end{figure}

\begin{figure}
    \centering
    \begin{subfigure}{0.4\textwidth}
    \centering
        \includegraphics[scale=0.18]{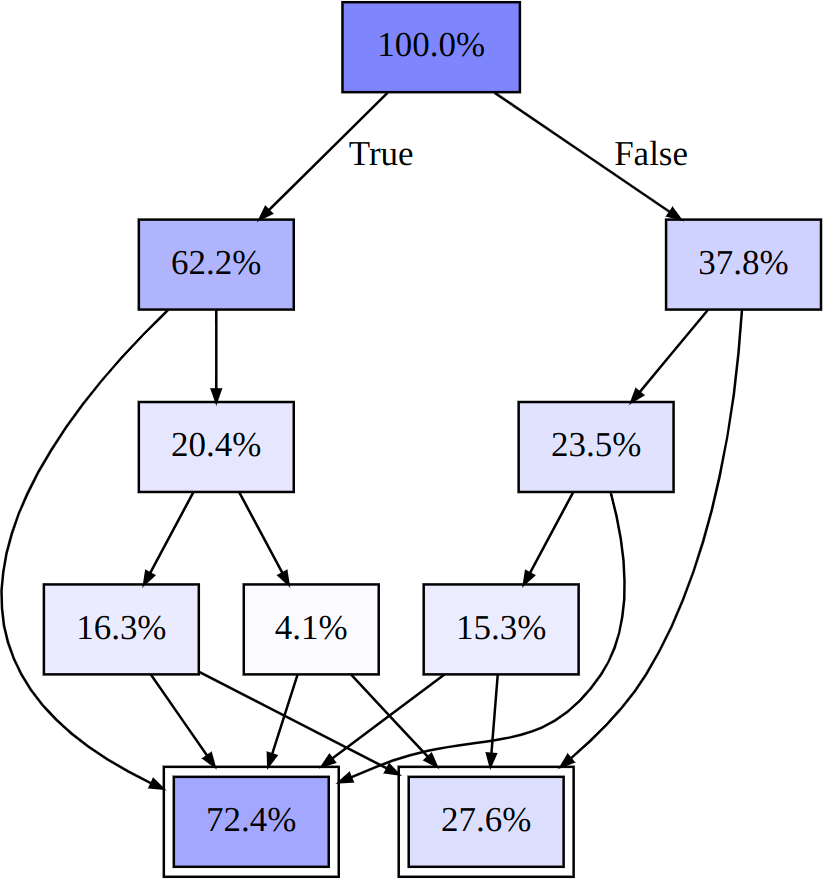}
        \caption{ODT structure}
    \end{subfigure}
    ~
    \begin{subfigure}{0.4\textwidth}
    \centering
        \includegraphics[scale=0.18]{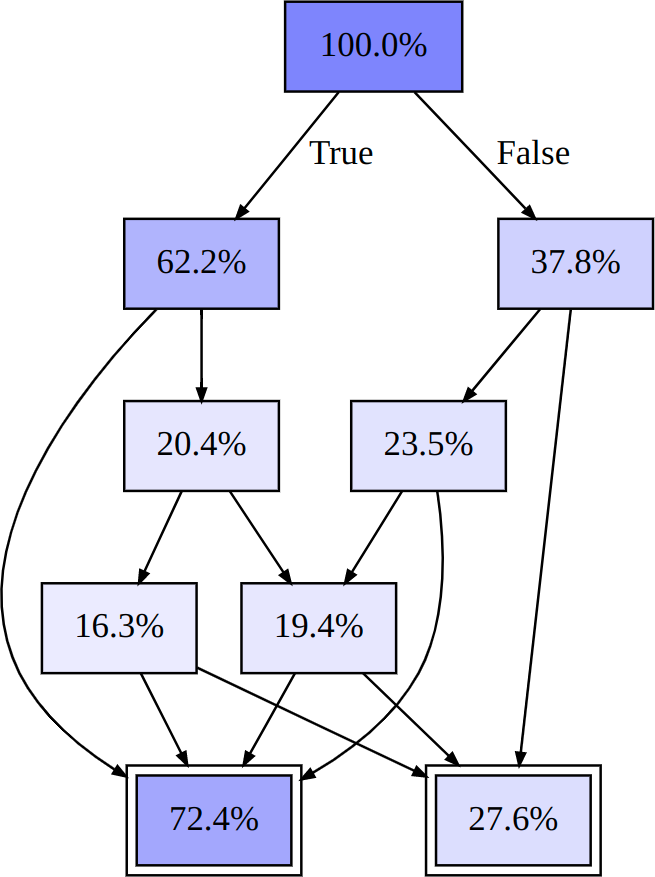}
        \caption{ODD structure}
    \end{subfigure}
    \caption{Fragmentation of an ODT and an ODD trained on the ``parkinsons'' dataset}
    \label{fig:frag3}
\end{figure}

\begin{figure}
    \centering
    \begin{subfigure}{0.4\textwidth}
    \centering
        \includegraphics[scale=0.18]{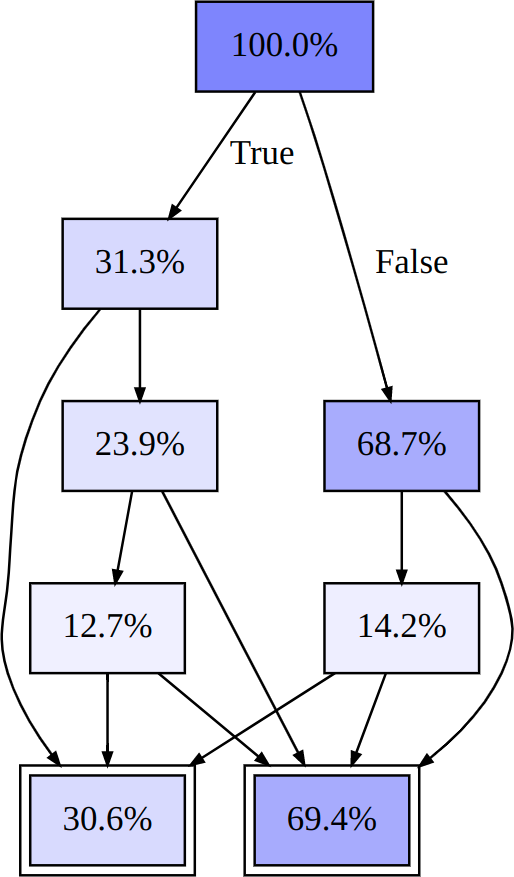}
        \caption{ODT structure}
    \end{subfigure}
    ~
    \begin{subfigure}{0.4\textwidth}
    \centering
        \includegraphics[scale=0.18]{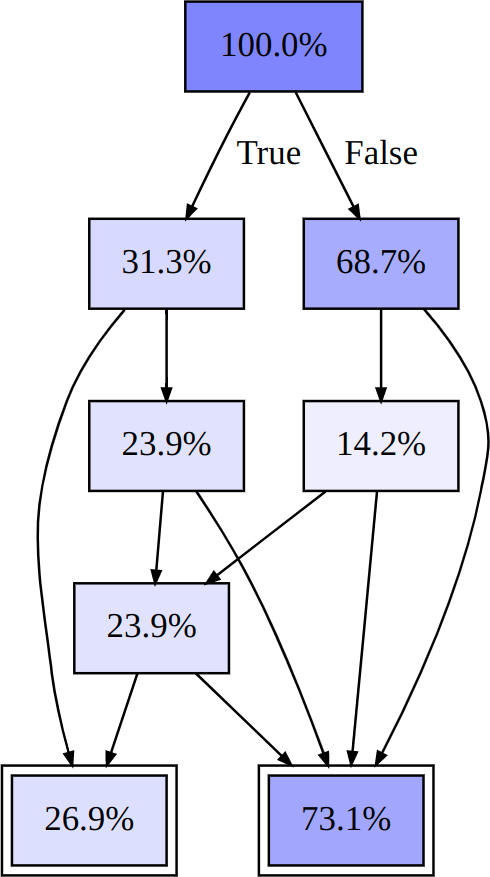}
        \caption{ODD structure}
    \end{subfigure}
    \caption{Fragmentation of an ODT and an ODD trained on the ``spect-heart'' dataset}
    \label{fig:frag4}
\end{figure}

\begin{figure}
    \centering
    \begin{subfigure}{0.4\textwidth}
    \centering
        \includegraphics[scale=0.18]{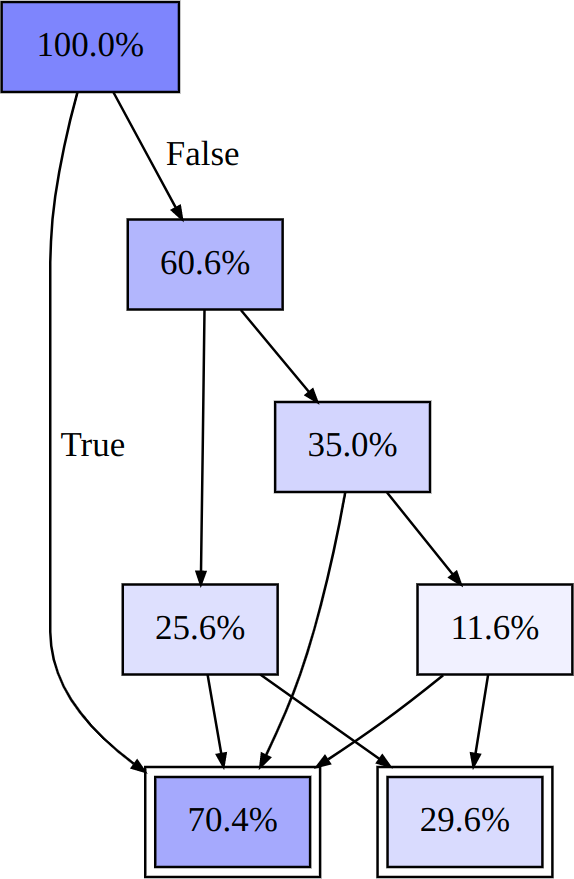}
        \caption{ODT structure}
    \end{subfigure}
    ~
    \begin{subfigure}{0.4\textwidth}
    \centering
        \includegraphics[scale=0.18]{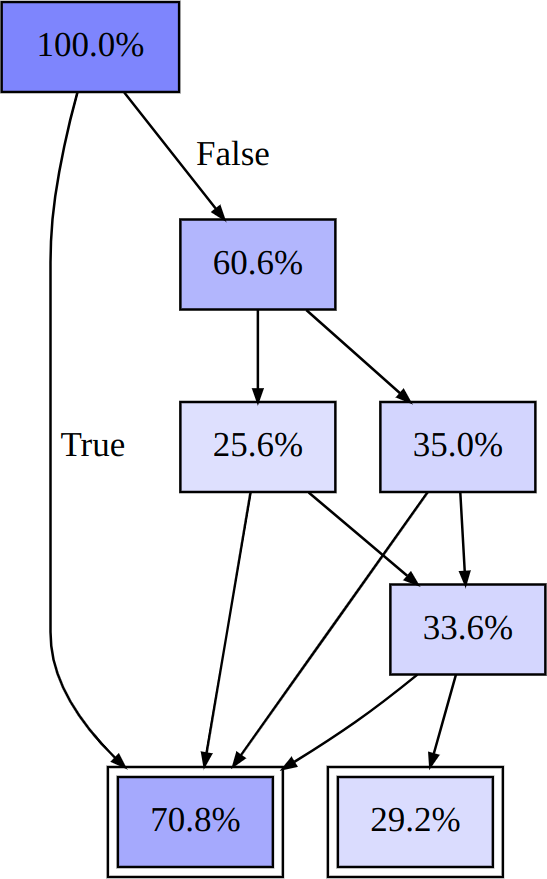}
        \caption{ODD structure}
    \end{subfigure}
    \caption{Fragmentation of an ODT and an ODD trained on the ``statlog german credit'' dataset}
    \label{fig:frag5}
\end{figure}

Low data fragmentation is a desirable characteristic of tree and diagram classifiers, as it improves model interpretability and stability when confronted with changing data. In general, ODDs are less prone to data fragmentation, as they allow nodes along an internal layer to consolidate sample flow from the layer immediately above. Figures \ref{fig:frag1}-\ref{fig:frag5} illustrate several datasets where ODDs lead to considerably less data fragmentation than ODTs, as observed in our experiments.

\end{document}